\pdfoutput=1

\documentclass[11pt]{article}

\usepackage[final]{acl}

\usepackage{times}
\usepackage{latexsym}

\usepackage[T1]{fontenc}

\usepackage[utf8]{inputenc}

\usepackage{microtype}

\usepackage{inconsolata}

\usepackage{graphicx}

\usepackage[many]{tcolorbox}
\usepackage{makecell} 
\usepackage{multirow} 
\usepackage{xcolor}
\usepackage{subfigure}
\usepackage{wrapfig}
\usepackage{enumitem}
\usepackage{amsthm}

\usepackage{amsmath}

\usepackage{amssymb}  
\usepackage{mathtools}

\usepackage{algorithm}   

\usepackage{algpseudocode}
\usepackage{adjustbox}
\usepackage{booktabs}

\definecolor{lightred}{rgb}{1, 0.88, 0.88}
\definecolor{lightgreen}{rgb}{0.88, 1, 0.88}
\definecolor{darkred}{rgb}{0.8, 0, 0} 
\definecolor{darkgreen}{rgb}{0, 0.6, 0} 

\newcommand{\up}[1]{\textsuperscript{\scriptsize \textcolor{darkred}{$\uparrow$#1}}}
\newcommand{\down}[1]{\textsuperscript{\scriptsize \textcolor{darkgreen}{$\downarrow$#1}}}

\theoremstyle{definition}
\newtheorem{definition}{Definition}

\newcommand{\argmin}{\operatornamewithlimits{arg\,min}}
\newcommand{\argmax}{\operatornamewithlimits{arg\,max}}

\title{Efficient Pretraining Data Selection for Language Models via Multi-Actor Collaboration}

\author{
 \textbf{Tianyi Bai\textsuperscript{1,2}},
 \textbf{Ling Yang\textsuperscript{3}},
 \textbf{Zhen Hao Wong\textsuperscript{3}},
  \textbf{Fupeng Sun\textsuperscript{5}},\\
 \textbf{Xinlin Zhuang\textsuperscript{2}},
 \textbf{Jiahui Peng\textsuperscript{2}},
 \textbf{Chi Zhang\textsuperscript{2}},
 \textbf{Lijun Wu \textsuperscript{2}},\\
  \textbf{Jiantao Qiu\textsuperscript{2}}\thanks{Corresponding authors.}\thanks{Project lead.},
 \textbf{Wentao Zhang\textsuperscript{2,3,4*}},
 \textbf{Binhang Yuan\textsuperscript{1*}},
 \textbf{Conghui He\textsuperscript{2*}},
\\
 \textsuperscript{1}Hong Kong University of Science and Technology,
 \textsuperscript{2}Shanghai Artificial Intelligence Laborator,\\
 \textsuperscript{3}Peking University,
\textsuperscript{4}Zhongguancun Academy,
 \textsuperscript{5}Imperial College London
\\
 \small{
   \textbf{Correspondence:} 
   {wentao.zhang@pku.edu.cn, biyuan@ust.hk, \{heconghui, qiujiantao\}@pjlab.org.cn}
 }
}

\begin{document}
\maketitle

\begin{abstract}

Efficient data selection is crucial to accelerate the pretraining of language model (LMs). While various methods have been proposed to enhance data efficiency, limited research has addressed the \textit{inherent conflicts} between these approaches to achieve optimal data selection for LM pretraining. 
To tackle this problem, we propose a \textit{multi-actor collaborative data selection} mechanism: each data selection method independently prioritizes data based on its criterion and updates its prioritization rules using the current state of the model, functioning as an independent actor for data selection; and a console is designed to adjust the impacts of different actors at various stages and dynamically integrate information from all actors throughout the LM pretraining process.
We conduct extensive empirical studies to evaluate our multi-actor framework. The experimental results demonstrate that our approach significantly improves data efficiency, accelerates convergence in LM pretraining, and achieves an average relative performance gain up to $10.5\%$ across multiple language model benchmarks compared to the state-of-the-art methods. 
Code and checkpoints are publicly released at 
{\url{https://github.com/Relaxed-System-Lab/multi-actor-data-selection}}.

\end{abstract}

\section{Introduction}

Efficient data selection is crucial for the pretraining of language model (LMs), as the quality of training data significantly impacts the statistical efficiency of the training procedure and the model performance~\citep{gpt3, glam, palm}. Recently, we have witnessed numerous approaches, such as filtering high-quality data~\citep{dsir, qurating}, mixing data from multiple domains~\citep{doremi, regmix}, and selecting data that optimally boosts downstream task performance dynamically~\citep{dsdm,mates}, which aim to improve data efficiency by prioritizing more informative training samples. However, these methods often operate independently or in isolated settings, limiting their potential when integrated into a collaborative framework. In this work, we want to explore \textit{how to effectively, flexibly, and robustly combine these advanced data selection techniques through the dynamic pretraining process}, addressing the challenges of optimizing data efficiency for LM pretraining at scale. 

\begin{figure}[t]
    \centering
    \includegraphics[width=0.42\textwidth]{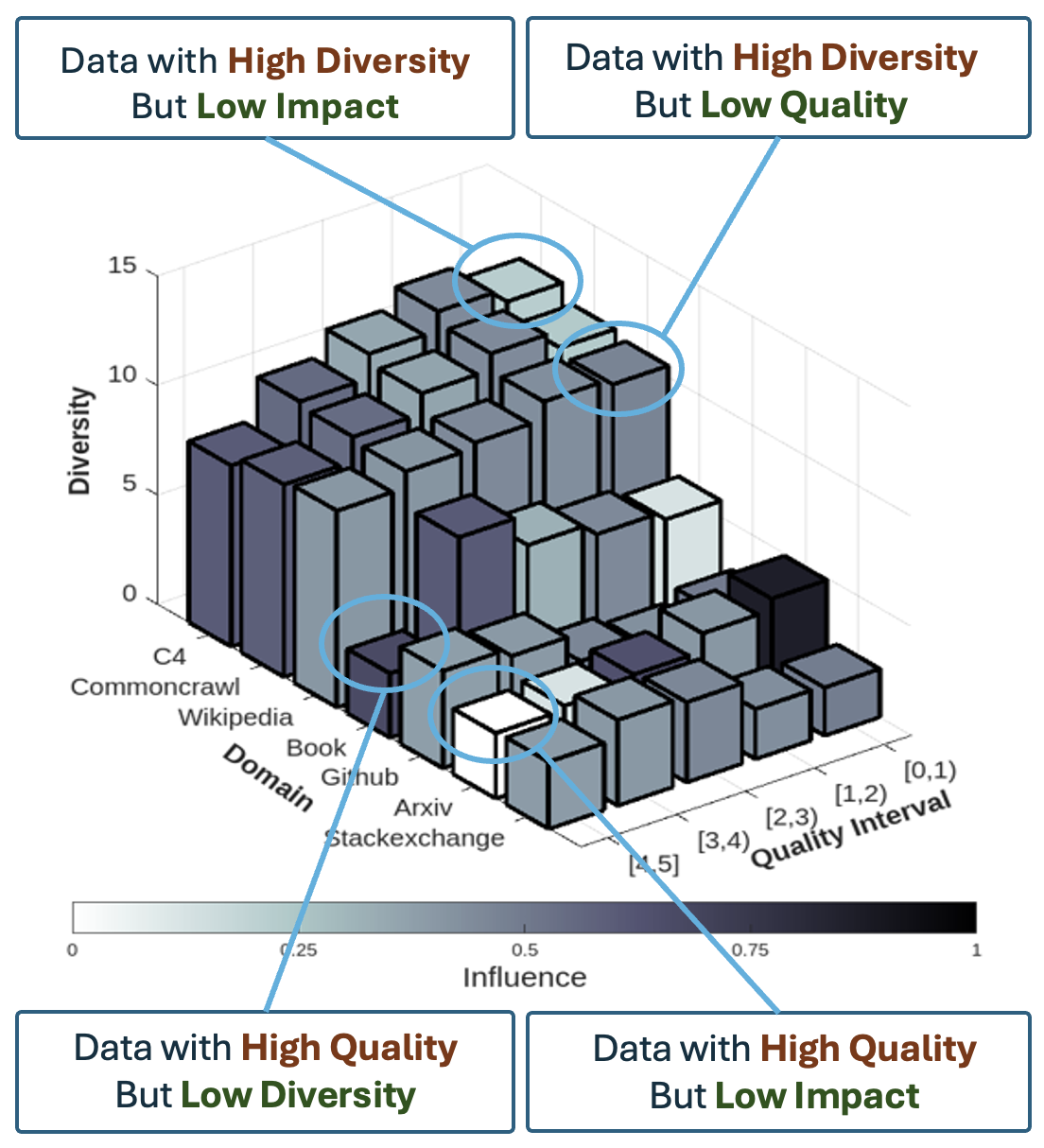}
    \caption{This figure shows the 627B token distribution by quality, domain, topic diversity, and model impact at step 1500. Each bar represents a subset, highlighting trade-offs between diversity, quality, and influence.}
    \label{fig:motivation}
    \vspace{-1.5em}
\end{figure}

Nowadays, various heuristic methods have been proposed to provide measurements for the data samples used during LM pre-training, aiming to optimize data efficiency by selecting or weighting the most informative training examples. However, we observe that integrating multiple data selection and mixing strategies presents significant challenges due to their \textit{inherent conflicts}. For example, high-quality data identified by scoring functions may not align with data that strongly impact model performance as measured by influence functions~\citep{dsdm}; similar conflicts also exists between other methods --- further details are enumerated in \S \ref{sec:case}. 
These observations actually motivate us to launch a systematic discussion about how to effectively integrate these methods during the dynamic pretraining process that provides superior data efficiency for LM pretraining.


Effectively integrating multiple data selection methods into a unified framework presents significant challenges. While each method may individually offer benefits, combining them requires navigating an exponential search space of possible configurations, which quickly becomes infeasible at scale. This challenge is further exacerbated in online data selection settings, where decisions must be made dynamically during training. Unlike offline approaches that rely on static, precomputed heuristics—such as classifier-based scoring, domain weighting, or topic sampling~\citep{gpt3,llama3,doremi}—online methods aim to adaptively select data based on the model’s evolving learning state. For example, MATES~\citep{mates} continuously probes the pretraining model to estimate the influence of individual data points and trains a lightweight data influence model to predict which samples will be most beneficial at each stage of training. Similarly, DSDM~\citep{dsdm} formulates data selection as an optimization problem, selecting data subsets that maximize performance on downstream tasks without relying on predefined quality metrics. Despite their effectiveness, these methods incur substantial computational overhead, as they require labeling or evaluating the entire dataset at each stage of training—an approach that is impractical for large-scale pretraining. Consequently, the core challenge is to develop an online integration framework that preserves the adaptivity and performance benefits of these techniques while remaining computationally efficient and scalable.

To address these challenges, we conduct a case study to identify the inherent conflicts for existing data selection methods and provide a multi-actor collaborative data selection framework to resolve this issue. Our multi-actor framework is inspired by the classical definition of intelligence outlined by \citep{russell2016artificial}, where an actor is defined as an entity that perceives a state and maps the observed state to actions. In our approach, data selection is achieved by the collaboration of these actors.
Our contributions can be summarized as:



\noindent\underline{\textbf{Contribution 1:}} Our case study on the SlimPajama reveals conflicts among four data selection metrics in LM pretraining. Despite these conflicts, prior studies~\citep{qurating,doremi,mates} show that even a single metric can effectively guide training, highlighting the need for better integration of these approaches.

\noindent\underline{\textbf{Contribution 2:}} 
We propose a novel multi-actor collaborative data selection framework (§\ref{sec:multi-actor}), where each method acts as an independent scorer for training data. An actor console integrates these scores to optimize selection, and a dynamic collaboration mechanism adjusts actor contributions throughout training, enhancing flexibility and data efficiency.

\noindent\underline{\textbf{Contribution 3:}} 
Extensive experiments demonstrate that: (\underline{1}) our method significantly improves data efficiency, accelerating LM convergence and achieving up to a 10.5\% performance boost over baselines (§\ref{sec:exp:e2e}); and (\underline{2}) ablation studies confirm that key components of our framework are essential for these gains (§\ref{sec:exp:ablation}). 

\vspace{-0.5em}


\section{Case Study - Inherent Conflicts in Data Selection}
\label{sec:case}
In this section, we present several observations derived from the SlimPajama datasets~\citep{slimpajama}, which reveal some inherent conflicts for different data selection measurements. 
To conduct this case study, we first label all data from the SlimPajama datasets using the quality scorer FineWeb-Edu~\citep{lozhkov2024fineweb-edu}. We then divide the data into subsets based on domain and quality ranges. From each subset, we uniformly sample data to assess topic diversity, i.e., the topic classification of the sampled data according to our methods. We analyze this diversity by examining the topic distribution within each subset. Additionally, we compute the normalized influence of the data on a pretrained 1.3B model at the 1500th step using influence functions to evaluate the data's impact on the model~\citep{dsdm}. Figure \ref{fig:motivation} illustrates the results, which presents a bar chart representing four dimensions: quality, domain, topic diversity, and influence on the pretrained model. The \textbf{x-axis} shows data quality, with higher intervals reflecting better scores from the FineWeb-Edu quality scorer. The \textbf{y-axis} indicates the dataset's domain, while the \textbf{z-axis} shows topic diversity within each subset, with taller bars indicating more diversity. The \textbf{color gradient} represents influence on the model, with darker shades showing greater impact. From this analysis, we highlight the following interesting observations:

\begin{itemize} [topsep=5pt, leftmargin=*]
\item \textit{High-quality data identified by the quality scorer may not have a significant impact on model performance.} For example, ArXiv documents rated between 4 and 5 by the scorer are considered high-quality. However, at the 1500th training step, they exert minimal impact on the model according to the influence functions, revealing a discrepancy between data quality and model impact. This observation is consistent with the previous discussion in \cite{dsdm}. 
\vspace{-1em}
\item \textit{High-quality data may exhibit low topic diversity.} Documents in the Book domain with a quality score of 4 to 5 are classified as high-quality by the scorer. Nevertheless, 85\% of these documents belong to the same topic, indicating a lack of diversity. 
\vspace{-1.2em}
\item \textit{Data with high topic diversity may not strongly influence model performance.} Documents from the C4 domain display considerable topic diversity. However, at the 1500th training step, they have limited impact on the model as measured by the influence functions, suggesting a conflict between diversity and model influence. 
\vspace{-1.2em}
\item \textit{Data with high topic diversity can be low quality.} Wikipedia documents show substantial topic diversity, which benefits the topic classifier. However, some of these documents are rated as low-quality by the quality classifier, revealing a trade-off between diversity and quality. 
\end{itemize}

We believe this inherent conflict illustrates that a naive ensemble of these mechanisms may lead to poor performance in terms of data efficiency for LM pretraining, which motivates the design and implementation of our multi-actor collaborative framework in \S \ref{sec:multi-actor}.  

\vspace{-0.5em}
\section{Collaborative Data Selection}
\label{sec:multi-actor}

In this section, we present the formalization of the data selection problem in \S \ref{sec:methods:formulation}, outline the overall framework of our methods in \S \ref{sec:methods:framework}, and detail the actor initialization and update in \S \ref{sec:methods:actor}, along with the collaborative mechanism in \S \ref{sec:methods:collaborate}.

\vspace{-0.5em}

\subsection{Problem Formulation}
\label{sec:methods:formulation}
We follow the definition of the data selection problem in \cite{dsdm} and \cite{mates} with slight modification. The objective for data selection is to choose a subset of size \( k \) from the entire pretraining dataset in such a way that the trained model's loss on downstream tasks is minimized. Let \( \mathcal{O} \) represent an optimization algorithm that maps a training dataset to a trained model. The optimal subset \( \mathcal{D}_k^* \) of the pretraining dataset \( \mathcal{D} \) can be expressed as:
\begin{equation}
\begin{small}
\begin{aligned}
    \mathcal{D}_k^* &\coloneqq \argmin_{\mathcal{D}_k \subset \mathcal{D}, |\mathcal{D}_k| = k} \mathcal{L}(\mathcal{D}_k\mid \mathcal{M},\mathcal{T}_{\text{eval}}),
    \label{eq:optimization}
\end{aligned}
\end{small}
\end{equation}
where 
\begin{equation*}
\begin{small}
\begin{aligned}
\mathcal{L}(\mathcal{D}_k\mid \mathcal{M},\mathcal{T}_{\text{eval}}) \coloneqq \mathbb{E}_{x \sim \mathcal{T}_{\text{eval}}} \left[\ell(x; \mathcal{O}(\mathcal{M},\mathcal{D}_k))\right]
\end{aligned}
\end{small}
\end{equation*}
denotes the expected loss (e.g., cross-entropy loss) for model $\mathcal{M}$ on downstream task $\mathcal{T}_{\text{eval}}$. 
Minimizing this objective directly is computationally challenging. Given that the real downstream tasks are unknown during model training, prior works have approximately optimized this problem by minimizing the loss on selected reference tasks $D_\text{ref}$ (e.g. LAMBADA~\cite{paperno2016lambada}, SQuAD~\cite{rajpurkar2016squad} and Jeopardy~\cite{tunguz2019jeopardy} in \cite{dsdm}). Specifically, they train proxy models to compute one-step training loss \citep{mates} or influence functions \citep{dsdm} on the reference tasks to approximate the true loss. However, this approach heavily depends on the selection of the reference tasks, while the chosen reference tasks may not be fully representative of all potential downstream tasks.

\begin{figure*}[tb]
    \centering
    \includegraphics[width=1\textwidth]{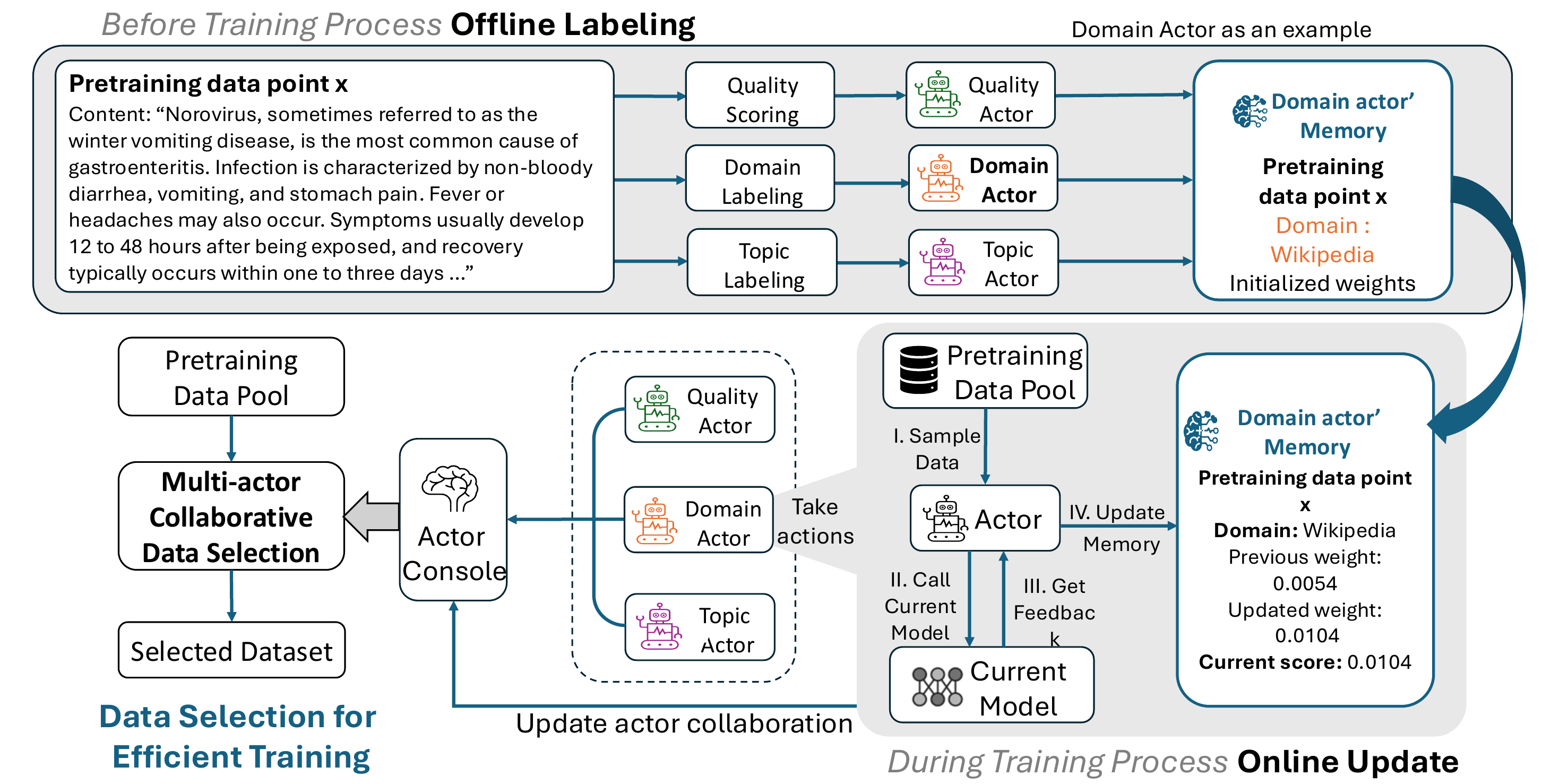}
    \caption{\textbf{Illustration of multi-actor collaborative framework.} Multi-actor collaborative framework for pretraining data selection that integrates multiple perspectives by combining offline priors and online model-derived preferences.}
    \label{fig:main}
    \vspace{-1.5em}
\end{figure*}

To avoid this obstacle, we do not directly minimize the loss on the reference tasks. Instead, we view this loss as a \textit{reward signal} that guides the update of predefined data selection methods. Concretely, we define a reward function $R(\mathcal{D}_k \mid \mathcal{M}, \mathcal{T}_{\text{ref}})$, where the reward is based on the performance gain of current model $\mathcal{M}$ trained on the subset $\mathcal{D}_k $ and evaluated on the reference tasks $\mathcal{T}_{\text{ref}}$. Then our optimization goal becomes maximizing this reward over time, as:
\begin{equation}
\begin{small}
\begin{aligned}
    &\mathcal{D}_k^* = \argmax_{\mathcal{D}_k \subset \mathcal{D}, |\mathcal{D}_k| = k} 
    \mathbb{E} \left[ R(\mathcal{D}_k \mid \mathcal{M}, \mathcal{T}_{\text{ref}}) \right]
    \label{eq:actor-optimization}
\end{aligned}
\end{small}
\end{equation}
where
\begin{equation*}
\begin{small}
\begin{aligned}
 R(\mathcal{D}_k | \mathcal{M}, \mathcal{T}_{\text{ref}}) \coloneqq \mathbb{E}_{x \sim \mathcal{T}_{\text{ref}}} \left[-\ell(x; \mathcal{O}(\mathcal{M},\mathcal{D}_k))\right].
\end{aligned}
\end{small}
\end{equation*}

We will rewrite $R(\mathcal{D}_k \mid \mathcal{M}, \mathcal{T}_{\text{ref}})$ as $R(\mathcal{D}_k)$ if there is no ambiguity.
In practice, $R(\mathcal{D}_k)$ can be approximated by the weighted average of influence functions~\cite{dsdm,mates}, which is defined by 
\begin{equation}
\begin{small}
\begin{aligned}
\label{eq:influence}
    r(x_i)
    = - \nabla_{\mathcal{M}}\mathcal{L}(\mathcal{T}_{\text{ref}}\mid \mathcal{M})^\top H^{-1}_{\mathcal{M}} \nabla_{\mathcal{M}} \mathcal{L}(x_i\mid{\mathcal{M}}),
\end{aligned}
\end{small}
\end{equation}
where $H_{\mathcal{M}}=\frac{1}{n}\sum_{i=1}^n\nabla^2_{\mathcal{M}} \mathcal{L}_{\mathcal{M}}(x_i\mid{\mathcal{M}})$ is the Hessian and its positive definite. Details of calculating influence functions for pretraining data point can be found in \S \ref{sec:app:theo}.

\subsection{Multi-Actor Data Selection Framework}
\label{sec:methods:framework}

In order to solve the optimization problem in Equation \ref{eq:actor-optimization}, we develop a framework illustrated in Figure \ref{fig:main}. This framework consists of two primary stages: the offline labeling stage and the online update stage. 
Before the training process, some initial information (i.e., the initialized measurements in some heuristic) is computed for the entire pretraining corpus,
and this information is stored separately in each actor's memory (formally defined below). During the training process, the current model (i.e., LM to be trained) is used to update the actors' memory and their collaboration mechanism based on rewards computed on the current model. An actor console is responsible for aggregating the opinions of each actor and making the final data selection decision. Formally, we define the actor in Definition \ref{def:actor} and the actor console in Definition \ref{def:actor_console}. Detailed formulation is in Appendix \ref{sec:app:marl}.

\begin{definition}[Actor]
\label{def:actor}
An \emph{actor} $ \mathcal{A}$ is a data selection method defined by a specific attribute (e.g., quality, domain, or topic) with memory $\mathcal{H}_{\mathcal{A}}$ that stores labels for each data point and their associated scores. 
During training, the actor takes several actions: (1) Sample data $\mathcal{D}_{\mathcal{A}}$ according to predefined sampling distribution, (2) Call the current model to compute the reward $\mathcal{R}(\mathcal{D}_{\mathcal{A}})$, (3) Get feedbacks from current model state, and (4) Update the internal weights $\mathbf{w_{\mathcal{A}}}$ in its memory. Then it assigns a score $\mathcal{S}_\mathcal{A}$ to each data point based on its updated memory, prioritizing the good data according to the updated weights.
One actor’s objective is to maximize its reward by updating this actor's internal weights and increasing the score of higher-reward data points.
\end{definition}

\begin{definition}[Actor Console]\label{def:actor_console}
The \emph{actor console} is in charge of coordinate opinions from different actor to make final decision of selecting dataset for next training stage.
Specifically, it consolidates scores $ \mathcal{S}_{\mathcal{A}}(x_i) $ from multiple actors $ \{\mathcal{A}_1, \dots, \mathcal{A}_n\} $ to calculate a final score $ S(x_i) $ for each data point $ x_i $, and select the final dataset. The console adjusts the collaborative weights $ \theta_{\mathcal{A}} $ for each actor based on their respective aggregate rewards $\mathcal{R}(\mathcal{D}_{\mathcal{A}})$, balancing their contributions during training. In cases where there are conflicts in the decisions made by actors, the console resolves these by adjusting the weights $ \theta_{\mathcal{A}} $ to prioritize the actors that have a greater positive impact on the model's performance, ensuring an effective data selection process.
\end{definition}

Now the reward signal is actually came from multiple actors, the optimization goal in Equation \ref{eq:actor-optimization} becomes maximizing the expectation of collaborative actors.
In our current implementation, we include three actors, which are topic actors, quality actors and domain actors. They are aiming to maximize the rewards from topic, quality and domain perspective respectively.
In the following sections, we will detail how we initialize and update a single actor (\S \ref{sec:methods:actor}),  and how we update the actor console for multi-actor collaboration (\S \ref{sec:methods:collaborate}).
\vspace{-0.5em}

\subsection{Single Actor Initialization and Update}
\label{sec:methods:actor}

\paragraph{Actor initialization.}
As defined in Definition \ref{def:actor}, for a particular actor, we have to maintain its memory $\mathcal{H}_\mathcal{A}$ throughout the training process. 
Before training process begin, we label the whole training dataset $\mathcal{D}$ offline and store the labeled information to the memory of corresponding actors. Specifically, for each data point $x_i \in \mathcal{D}$, i.e., a single document in our settings, we first get the quality, topic and domain label using scorer and classifier.

We initialize the weight of the topic actor and domain actor following the RegMix~\cite{regmix} framework. Unlike the original RegMix, which only considers mixing data based on domain labels, we examine data mixing weights based on domain as well as the topic labels. We initialize our quality actor similar to the data selection decision of QuRating~\cite{qurating} and FineWeb-Edu~\cite{lozhkov2024fineweb-edu}. Further details can be found in \S \ref{sec:app:weight_initialize}. The initial weights for each actor are stored in their respective memory.

During the training phase, we leverage the current model to adjust the weight of each actor. As depicted in Figure \ref{fig:main}, at the data selection stage, each actor performs several actions to update its memory and inform decision-making. Take domain actor as an example, it takes three-step action during data selection stage: (\underline{1}) Sample pretraining data points from the pretraining data pool, distributing them uniformly across each domain; (\underline{2}) Call the current model to assess the reward of each data point and gather feedback; (\underline{3}) Update the memory of domain weights based on gathered feedback and adjusts the score for each data point by incorporating prior knowledge from the offline labeling process. This process is similarly followed by the quality actor and the topic actor. 

\paragraph{Actor update.} 
After sampling data uniformly from actor search space, each actor updates its internal weights using local information based on the sampled data. For example, for domain actor, it calculates the average influence of each domain.
For actor $\mathcal{A}\in\{\mathcal{A}_\text{Quality}, \mathcal{A}_\text{Domain}, \mathcal{A}_\text{Topic}\}$, it updates its internal weights by calculating the overall rewards sampled from each interval as:
\begin{equation}
\begin{small}
\vspace{-0.5em}
\label{eq:agent_score}
\mathcal{R}(\mathcal{D}_{\mathcal{A}}^j)= \frac{1}{|D_{\mathcal{A}}^j|} \sum_{x_i \in D_{\mathcal{A}}^j} r(x_i)\coloneqq
\overline{R}_{\mathcal{A}}^j ,
\end{small}
\vspace{-0.5em}
\end{equation}
where \(D_{\mathcal{A}}^j\) represents the sample set of the \(j\)-th subcategory under actor $\mathcal{A}$, e.g. Wikipedia for domain actor. And \(x_i\) is a sample within this sample set. 
The sliding averaging is used to update the weight for each subcategory $w_{\mathcal{A}}^j$ with current rewards:
\vspace{-0.5em}
\begin{equation}
\begin{small}
\label{eq:agent_update}
    w_{\mathcal{A}}^j \leftarrow (1-\eta_{A}) \cdot w_{\mathcal{A}}^j + \eta_{A} \cdot \overline{R}_{\mathcal{A}}^j,
\end{small}
\vspace{-0.5em}
\end{equation}
where $\eta_{\mathcal{A}}$ is the sliding average factor to tradesoff bias-variance. The overall updated weight of actor $\mathcal{A}$ is an vector in $n_\mathcal{A}$ dimension, $\mathbf{w}_{\mathcal{A}} = [w_{\mathcal{A}}^1, ..., w_{\mathcal{A}}^{n_\mathcal{A}}]$, where $n_\mathcal{A}$ is the number of total subcategory within the space of actor $\mathcal{A}$. Utilizing the prior memory stored by each actor, it can give out a final score for each data point as $\mathcal{S}_\mathcal{A}(x_i)=w_{\mathcal{A}}^j$, where $j$ is the subcategory that $x_i$ belongs to.
\subsection{Multi-Actor Collaboration}
\label{sec:methods:collaborate}

Ultimately, the actor console defined in Definition \ref{def:actor_console} aggregates all actors' feedback to compute a final score for each data point, determining the final data selection decision. 

\paragraph{Multi-actor collaboration.}
In the context of multi-actor collaboration, the weighted score for each actor must be calculated to evaluate their respective contributions effectively. This calculation takes into account various factors specific to each actor. For every data sample \(x_i\), the overall score \(S(x_i)\) is determined by the following formula:
\begin{equation}
\begin{aligned}
\label{eq:coordinator}
    &S(x_i) = (\theta_{\text{Quality}} \cdot \mathcal{S}_{\text{Quality}}(x_i) + \\
    &\theta_{\text{Domain}} \cdot \mathcal{S}_{\text{Domain}}(x_i)\
    + \theta_{\text{Topic}} \cdot \mathcal{S}_{\text{Topic}}(x_i)),
\end{aligned}
\end{equation}
where $\mathcal{S}_{\text{Quality}}(x_i)$, $\mathcal{S}_{\text{Domain}}(x_i)$, and $\mathcal{S}_{\text{Topic}}(x_i)$ are scores calculated by quality, domain, and topic actors for sample $x_i$. $\mathbf{\theta_\mathcal{A}}\in\{\theta_{\text{Quality}},\theta_{\text{Domain}}, \theta_{\text{Topic}}\}$ is the collaborative weight for each actor, which is updated during training process.

\begin{table*}[tb]
\centering
\caption{Our approach boosts model performance across tasks. To fit demonstrations within 1024 tokens, we provide full results for 0, 3, and 5 shots in Appendix~\ref{sec:app:full_results}. The table covers Problem Solving, Commonsense Reasoning, and Reading Comprehension, with best QuRating and DSIR variants: QuRating-Edu and DSIR-Wiki.}\label{table:main}
\vspace{-0.75em}
\adjustbox{max width=0.9\textwidth}{
\begin{tabular}{l>{\centering\arraybackslash}m{0.175\textwidth}>{\centering\arraybackslash}m{0.175\textwidth}>{\centering\arraybackslash}m{0.175\textwidth}>{\centering\arraybackslash}m{0.175\textwidth}}
\toprule
\multirow{2}{*}{\textbf{Selection Method}} & \textbf{Problem Solving} & \textbf{Commonsense Reasoning} & \textbf{Reading \ \ \ Comprehension} & \textbf{Average} \\
& \textit{(4 tasks)} & \textit{(4 tasks)} & \textit{(2 tasks)} & \textit{(10 tasks)} \\
\midrule
\textbf{Random sampling} - 30B tokens & 31.1 & 32.9 & 43.1 & 34.2 \\
\textbf{Random sampling} - 60B tokens & 33.6\up{2.5} & 33.7\up{0.8} & 46.1\up{3.0} & 36.1\up{1.9} \\
\midrule
\textbf{Perplexity} PPL~\citep{ppl} & 29.9\down{1.2} & 30.5\down{2.4} & 42.4\down{0.7} & 32.7\down{1.5} \\
\midrule
\textbf{Classifier-based data selection} & & & & \\
QuRating~\citep{qurating} & 34.1\up{3.0} & 34.1\up{1.2} & 41.4\down{1.7} & 35.6\up{1.4} \\
FineWeb-Edu~\citep{fineweb} & 32.6\up{1.5} & 33.0\up{0.1} & 45.3\up{2.2} & 35.3\up{1.1} \\
DSIR~\citep{dsir} & 30.9\down{0.2} & 32.0\down{0.8} & 41.5\down{1.6} & 33.5\down{0.7} \\
\midrule
\textbf{Domain mixing methods} & & & & \\
DOGE~\cite{doge} & 30.9\down{0.2} & 32.2\down{0.7} & 45.1\up{2.0} & 34.3\up{0.1} \\
DoReMi~\citep{doremi} & 30.4\down{0.7} & 32.6\down{0.3} & 44.8\up{1.7} & 34.1\down{0.1} \\
DMLaw~\citep{dmlaw} & 30.2\down{0.9} & 32.1\down{0.9} & 45.1\up{2.0} & 33.9\down{0.3} \\
RegMix~\citep{regmix} & 30.7\down{0.4} & 32.5\down{0.4} & 44.6\up{1.5} & 34.2\up{0.0} \\
\midrule
\textbf{Influence} MATES~\citep{mates} & 30.9\down{0.2} & 34.0\up{1.1} & 46.5\up{3.4} & 35.3\up{1.1} \\
\midrule
\textbf{Multi-actor collaboration (ours)} & 36.7\up{5.6} & 34.8\up{1.9} & 45.9\up{2.8} & 37.8\up{3.6} \\
\bottomrule
\end{tabular}
}
\vspace{-0.75em}
\end{table*}

\paragraph{Collaborative weight update.} 
To dynamically adjust the importance of each actor during various training phases, we modify the actor's collaborative weight based on its overall rewards. We compute the reward of each actor and the average reward across all actors:
\begin{equation}
\begin{small}
\label{eq:collaborative_score}
    \overline{R}_{\mathcal{A}} = \frac{1}{|n|} \sum_{j=1}^n w_{\mathcal{A}}^j \cdot \overline{R}_{\mathcal{A}}^j, \quad
    \overline{R} = \frac{1}{3} \sum_{\mathcal{A}} \overline{R}_{\mathcal{A}},
\end{small}
\end{equation}
This information is then used to update each actor’s collaborative weight, which is stored in the actor console’s memory for future decision-making:
\begin{equation}
\begin{small}
\label{eq:collaborative_update}
    \theta_\mathcal{A} \leftarrow \theta_\mathcal{A} + \eta_\mathcal{A} \cdot (\overline{R}_{\mathcal{A}} - \overline{R}).
\end{small}
\end{equation}
By continuously refining these weights, the collaboration strategy adapts to optimize overall performance and appropriately adjust the role of each actor throughout different stages of training.
Complete training pipeline is outlined in Algorithm \ref{algo:multi_agent}.

\section{Experiments}

We conduct a series of experiments to evaluate the effectiveness of our multi-actor collaborative data selection method. Comprehensively, we find that: (\underline{1}) In the end-to-end experiments, our approach introduces significant improvement in terms of data efficiency leading to faster convergence for LM training, and achieves up to $10.5\%$ improvements on average across various language model benchmarks when compared with other baseline approaches (\S \ref{sec:exp:e2e}); (\underline{2}) We also verify that the design and implementation of the core components in our multi-actor framework design are necessary to reach this advanced performance through a set of carefully designed ablation studies (\S \ref{sec:exp:ablation}).

\subsection{End-to-end Experiments}
\label{sec:exp:e2e}

We evaluate our multi-actor framework against a wide category of state-of-the-art approaches to compare the data efficiency for LM pretraining. We train a $1.3$ billion parameter \textsc{Llama-2} architecture model with $30$ billion selected tokens. We also report our results of generalizing our methods to \textbf{3.6B} and \textbf{8B} models in Appendix \ref{app:generalization}.

\noindent\textbf{Experimental setup.} We first enumerate the experimental setup as below:


\begin{itemize}[topsep=5pt, leftmargin=*]
\item \textbf{Pretraining datasets.}  We utilize the popular SlimPajama~\citep{slimpajama} dataset including $627$ billion tokens, which is derived from the RedPajama~\citep{together2023redpajama} dataset. The SlimPajama~\citep{slimpajama} provide the meta-data about the domain information for each sample. Before the training process, we annotate the entire dataset using the FineWeb-Edu quality scorer~\citep{fineweb} along with our custom-trained BERT-based topic classifier. The training details for the topic classifier is provided in Appendix \S \ref{sec:app:topic}.
\vspace{-0.35em}
\item \textbf{Training details.} We adopt the model architecture from \textsc{Llama-2}~\citep{touvron2023llama} at the scale of $1.3$ billion parameters (see the detailed configuration in Appendix \S \ref{sec:app:pretrain}-Table \ref{tab:model_architecture}). Following the principles of the scaling law~\citep{hoffmann2022training} and the DCLM framework~\citep{li2024datacomp}, we decide to use a total of 30 billion tokens. All training tokens are sampled from the 670 billion-token SlimPajama~\citep{slimpajama} dataset using various sampling strategies. Further details regarding the training process can be found in \S \ref{sec:app:pretrain}.

\vspace{-0.35em}
\item \textbf{Evaluation benchmarks.} To evaluate the pre-trained models thoroughly, we conduct extensive assessments across various downstream tasks, categorized into three areas: (\underline{1}) problem solving: MMLU~\citep{mmlu}, ARC-Easy/Challenge~\citep{arc}, and MathQA~\cite{sciq}; (\underline{2}) commonsense reasoning: SIQA~\citep{siqa}, WinoGrande~\citep{winogrande}, OpenbookQA~\citep{openbookqa}, and CommonsenseQA~\citep{commonsenseqa}; (\underline{3}) reading comprehension: RACE~\citep{race} and BoolQ~\citep{boolq}. Evaluations are conducted using the \texttt{lm-evaluation-harness} framework~\citep{eval-harness} in an in-context learning setting, and average accuracy is reported for easy comparison.

\vspace{-0.35em}
\item \textbf{Baselines.} We select a wide range of baselines to conduct extensive the data efficiency comparison, where these methods can be classified to five main categories: (\underline{1}) \textit{random sampling}, we test this policy with both the standard data volume of 30B tokens and a supplemented version with 60B tokens; (\underline{2}) \textit{perplexity}-based data selection~\cite{ppl}; (\underline{3}) \textit{classifier}-based data selection, where we select the following methods: QuRating~\cite{qurating}, FineWeb-Edu~\cite{fineweb}, DSIR-Book~\cite{dsir} and DSIR-Wiki~\cite{dsir}; (\underline{4}) \textit{domain mixing}-based methods, where we select the following methods: DOGE~\cite{doge}, DoReMi~\cite{doremi}, DMLaw~\cite{dmlaw} and RegMix~\cite{regmix}; and (\underline{5}) \textit{influence function} based methods for online data selection, i.e., MATES~\cite{mates}. Implementation details of these baselines can be found in \S \ref{sec:app:baseline}.

\end{itemize}

\noindent\textbf{Results.} 
We present the results of three types of downstream tasks in Table \ref{table:main}, with the complete 0-shot (Table \ref{table:zershot}), 3-shot (Table \ref{table:threeshot}), and 5-shot (Table \ref{table:fiveshot}) results for all tasks enumerated in \S \ref{sec:app:full_results}. 
We highlight that \textit{our methods show a substantial improvement in the average performance across all downstream tasks when compared with all the baselines}. Concretely, we observe that when compared with the \textit{random sampling} based approach, our method not only significantly outperforms the standard $30$ billion token setup but also surpasses the model trained on $60$ billion tokens with a performance gain of 4.7\%. Similarly, we also show an improvement of 15.6\% compared with \textit{perplexity}-based data selection~\cite{ppl}, an improvement of up to 6.2\% compared with \textit{classifier}-based data selection, an improvement of up to 10.2\% compared with \textit{domain mixing}-based methods, and an improvement of 7.1\% compared with \textit{influence function} based approach, i.e., MATES~\cite{mates}.

\begin{table*}[tb]
\centering
\caption{This ablation study examines the performance of various combinations of actor collaboration and update mechanisms. All models are in 1.3B LLaMA2 architecture. Three-shot accuracy is reported for all tasks, with the highest value in each column shown in \textbf{bold}. 
}\label{table:ablation_1.3B}
\vspace{-0.5em}
\adjustbox{max width=1\textwidth}{
\begin{tabular}{l>{\centering\arraybackslash}m{0.085\textwidth}>{\centering\arraybackslash}m{0.085\textwidth}>{\centering\arraybackslash}m{0.085\textwidth}>{\centering\arraybackslash}m{0.085\textwidth}>{\centering\arraybackslash}m{0.085\textwidth}>{\centering\arraybackslash}m{0.085\textwidth}>{\centering\arraybackslash}m{0.085\textwidth}>{\centering\arraybackslash}m{0.085\textwidth}>{\centering\arraybackslash}m{0.085\textwidth}>{\centering\arraybackslash}m{0.085\textwidth}>{\centering\arraybackslash}m{0.085\textwidth}}
\toprule
 & \multicolumn{4}{c}{ Problem Solving }  & \multicolumn{4}{c}{ Commonsense Reasoning} & \multicolumn{2}{c}{ Reading Compreh.} & \\
\cmidrule(lr){2-5} \cmidrule(lr){6-9} \cmidrule(lr){10-11} 
 \multirow{1}{*}{\textbf{Selection Method}}& \textbf{ARC-E} & \textbf{ARC-C} & \textbf{MathQA} & \textbf{MMLU} & \textbf{O.B.QA} & \textbf{SIQA} & \textbf{W.G.} & \textbf{C.S.QA} & \textbf{BoolQ} & \textbf{RACE} & \textbf{Average} \\
\midrule
Quality\&Domain\&Topic actor & \textbf{65.8} & \textbf{31.5} & \textbf{23.0} & 26.6 & \textbf{24.6} & 39.9 & \textbf{54.1} & 20.1 & \textbf{60.4} & 30.5 & \textbf{37.7} \\
without collaboration update & 59.4 & 26.3 & 21.3 & 25.1  & 20.5  & 38.9 &  52.9 & 19.8 & 58.1 & 28.3  & 35.1 \\
\midrule
Domain\&Quality actor & 63.3 & 29.7 & 22.6 & 25.1 & 21.8 & \textbf{40.5} & 53.1 & 20.3 & 59.5 & 28.8 & 36.5 \\
Topic\&Quality actor & 62.9 & 28.1 & 22.3 & 26.5 & 22.6 & 39.6 & 51.8 & \textbf{21.7} & 56.7 & \textbf{30.7} & 36.3 \\
Domain\&Topic actor & 55.6 & 25.2 & 21.8 & 26.5 & 23.1 & 39.1 & 53.7 & 20.9 & 57.5 & 29.0 & 35.2 \\
\midrule
Quality actor & 59.1 & 29.7 & 22.4 & 25.3 & 21.1 & 38.5 & 51.2 & 19.1 & 57.2 & 28.3 & 35.2 \\
Domain actor & 54.1 & 25.6 & 21.4 & 25.9 & 22.3 & 38.1 & 53.6 & 20.0 & 58.1 & 27.9 & 34.7 \\
Topic actor & 55.3 & 25.3 & 21.9 & \textbf{27.1} & 22.1 & 39.4 & 51.5 & 19.8 & 56.3 & 28.9 & 34.8 \\
\midrule
No actor & 54.6 & 23.0 & 22.1 & 24.9 & 18.8 & 40.3 & 52.9 & 21.5 & 53.0 & 29.8 & 34.1 \\
\bottomrule
\end{tabular}
}
\vspace{-0.75em}
\end{table*}

\begin{figure}
    \centering
    \vspace{-5pt} 
    \includegraphics[width=0.9\linewidth]{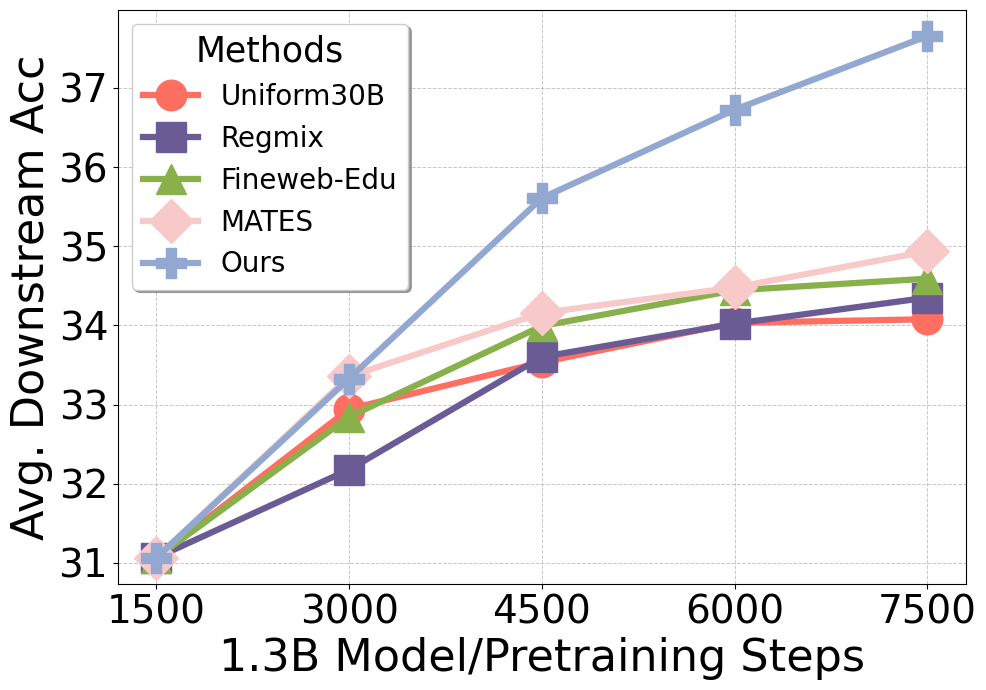}
    \caption{Downstream three-shot performance of the 1.3B model in relation to pretraining steps, using 7500 steps for 30B tokens. Our methods outperform baselines from all categories.}
    \label{fig:steps}
    \vspace{-1.5em}
\end{figure}

\noindent\textbf{Discussion.} We highlight that our proposed multi-actor collaborative data selection mechanism introduces statistical efficiency in terms of LM training convergence and also provides some computational efficiency in terms of data processing overheads. 
In terms of \textit{statistical efficiency}, our method consistently outperforms others at every benchmarked training step, as shown in Figure \ref{fig:steps}. While MATES~\cite{mates} performs comparably to our methods during the early training phase (steps $1500$ to $3000$), its performance drops in later stages. This aligns with its original paper, which notes that relying solely on influence functions for specific reference tasks (e.g., LAMBADA~\citep{paperno2016lambada}) can degrade performance in mid-to-late pretraining. Despite this, MATES still outperforms other methods without dynamic adjustments shown in Figure \ref{fig:steps}. In contrast, our multi-actor collaborative data selection mechanism can dynamically adjust the corresponding weights from different actors and select data based on the most up-to-date model preferences, effectively mitigating biases and surpassing other domain-mixing and data-selection techniques.
In terms of \textit{computational efficiency}, we also achieve higher computational efficiency than previous methods. For example, QuRating~\cite{qurating} requires around $7.13 \times 10^{20}$ FLOPs to label the entire SlimPajama dataset, while our offline labeling takes just $9.91 \times 10^{19}$ FLOPs. MATES~\cite{mates}, which recalculates influence scores and trains a BERT model for each labeling cycle, incurs $1.98 \times 10^{20}$ FLOPs for a four-stage update. Additionally, MATES’ labels are only usable in the next training stage, making it time-consuming and difficult to scale. In contrast, our method can improve the computational efficiency from two aspects: (\underline{1}) we find that a group of light-weight actors collaboratively enables superior data selection, which is more computational efficiency than any method that requires a heavy data processing or label procedure; (\underline{2}) the collaborative, dynamic learning procedure introduced in our multi-actor framework is computational efficient; by using a sampled holdout set and CPU-based calculations for updating actor parameters, our computational overhead is ignorable compared with heavy LM training computation.

\subsection{Ablation Study}
\label{sec:exp:ablation}

We introduce a set of carefully designed ablation studies to justify the design and implementation of our multi-actor collaborative data selection framework. Concretely, (\underline{1}) we test the combination of different actors to show the advance introduced by collaboration, and (\underline{2}) we verify the necessity of the dynamic adjustment of the actor's weight for data selection.

\noindent \textbf{Results and discussion.}
The results of the ablation study are shown in Table \ref{table:ablation_1.3B}. We want to highlight the result from two aspects:
\underline{First}, the ablation study underscores \textit{the importance of each actor in achieving optimal performance} across the training tasks. When the quality, domain, and topic actors are used together, the model performs best, highlighting the benefits of their combined use, as shown in Table \ref{table:ablation_1.3B}. In terms of evaluating \textit{each actors' contributions}, we find that the quality actor excels in problem-solving tasks like ARC-E and MathQA by leveraging educational knowledge but is less effective for domain-specific or context-heavy tasks like BoolQ and RACE. The domain actor enhances commonsense reasoning (e.g., CommonsenseQA) and reading comprehension (e.g., BoolQ) by incorporating domain-specific knowledge. The topic actor is most effective for multi-topic tasks like MMLU and contributes significantly to commonsense reasoning tasks like SocialIQA.
\underline{Second}, the ablation study verifies the design and implementation of \textit{the collaborative dynamic adjustment of the actors' weights} (introduced \S \ref{sec:methods:collaborate}) for efficient data selection. When actors were initialized with equal, fixed weights instead of using dynamic weighting, overall performance dropped significantly, as shown in Table~\ref{table:ablation_1.3B}. 

\section{Related Work}

\noindent \textbf{Data selection in LM pretraining.} Selecting high-quality pretraining data from large corpora is crucial for effective LM training. Recent approaches leverage various methodologies for efficient data selection. Concretely, \textit{classifiers}~\citep{gpt3,palm,glam,dsir} and \textit{language modeling perplexity}~\citep{ccnet,perplexity} have been applied to identify data resembling high-quality samples; recently, more advanced quality scores based on classifier have shown the effectiveness in data selection, e.g., QuRating~\citep{qurating}, FineWeb-Edu~\citep{lozhkov2024fineweb-edu}, etc. \textit{Data mixture} is another effective way to improve data diversity, at both token level~\citep{llama, pile, slimpajama} and sample level, e.g., DoReMi~\citep{doremi}, DOGE~\citep{doge}, DMLaw~\citep{dmlaw}, and RegMix~\citep{regmix}; very recently, topic distributions has also been considered as an effective data mixing method, e.g., the downsampling overrepresented topics in Llama 3.1~\citep{llama3}.
\textit{Influence functions} have been studied to understand for data efficiency ~\citep{influence2017}, and some recent attempts based on efficient approximation have been proposed to improve data efficiency in LM pretraining~\citep{scale-influence,lm-influence,isonuma2024unlearning};
For example, MATES~\citep{mates} uses a staged BERT model to assess data influence, QUAD~\citep{quad} leverages cluster information to reduce the computational cost of calculating individual data influence.

\noindent \textbf{Multi-agent and multi-actor collaborative frameworks.}
Collaboration across multiple autonomous or semi-autonomous entities has been studied extensively under the paradigms of multi-agent and multi-actor systems. Traditional multi-agent systems~\citep{russell2016artificial, wooldridge2009introduction} involve autonomous agents that make independent decisions and learn through interaction, often coordinated via mechanisms such as reward signals, negotiation, or centralized planning. These systems have been effective in applications such as neural architecture search~\citep{bello2017neural}, collaborative large language model programming~\citep{hong2024metagpt}, and distributed control~\citep{olfati2006flocking}.
In contrast, our work adopts a multi-actor perspective, where actors are not general-purpose intelligent agents, but specialized components—specifically, independently operating data selection methods. Each actor follows a distinct heuristic to prioritize pretraining data, adapting dynamically to the model’s evolving state. A central coordination mechanism integrates these priorities over time, resolving conflicts and guiding collective behavior. This aligns with formalizations of multiactor systems, where the term "actor" serves as a generic abstraction that encompasses agents, bodies, or effectors~\citep{russell2016artificial}. Unlike multi-agent systems, which may involve deliberative autonomy and decentralized goal negotiation, multiactor systems often assume a shared objective and focus on concurrent action, coordination, and conflict resolution.
Our framework draws from both traditions: it retains the modularity and adaptability of multi-agent systems, while leveraging the structured concurrency and coordination principles emphasized in multiactor planning~\citep{ligtenberg2001multi, ligtenberg2004design,wang2020multi}. This hybrid approach enables us to tackle a largely underexplored challenge: resolving competition among heterogeneous data selection heuristics to improve the efficiency and effectiveness of language model pretraining.
\section{Conclusions}
In this paper, we introduce a multi-actor collaborative data selection framework to enhance efficiency in LM pretraining. Our framework allows multiple data selection methods to operate as independent actors, with an actor console designed to dynamically integrate their outputs throughout the LM training process. Empirical studies show it improves data efficiency, speeds up convergence, and achieves up to 10.5\% average performance gains over state-of-the-art methods. These results demonstrate the effectiveness of dynamically combining data selection strategies to resolve conflicts and optimize LM pretraining.

\section*{Limitations}  
While our method greatly improves data selection for language model pre-training, our study has some limitations. Due to computational constraints, our experiments were limited to relatively small-scale models (up to 8B parameters) with restricted token budgets. Additionally, while our quality metrics are comprehensive, they may not fully capture all dimensions of pre-training data. In future work, we plan to refine or expand these metrics to bridge these gaps.

\section*{Acknowledgement}
This work is supported by the HKUST startup grant R9895 from CSE, RGC-ECS project 26218024, Shanghai Artificial Intelligence Laboratory, the National Key R\&D Program of China (2022ZD0160201), the National Key R\&D Program of China (2024YFA1014003), National Natural Science Foundation of China (92470121, 62402016), and CAAI-Ant Group Research Fund.
We express sincere thanks to InternTrain Team of Shanghai Artificial Intelligence Laboratory, especially Yang Gao, for their kind help for the pre-training experiments.

\newpage
\bibliography{custom}

\newpage
\appendix

\section{Appendix}
\begin{table*}[tb]
\centering
\caption{This study compares the performance of training \textbf{3B model} from scratch on 60B tokens selected using random sampling versus multi-actor collaboration. Accuracy is reported for all tasks, with the highest value in each column shown in \textbf{bold}.}
\label{table:ablation_3B}
\vspace{1em}
\adjustbox{max width=1\textwidth}{
\begin{tabular}{l>{\centering\arraybackslash}m{0.085\textwidth}>{\centering\arraybackslash}m{0.085\textwidth}>{\centering\arraybackslash}m{0.085\textwidth}>{\centering\arraybackslash}m{0.085\textwidth}>{\centering\arraybackslash}m{0.085\textwidth}>{\centering\arraybackslash}m{0.085\textwidth}>{\centering\arraybackslash}m{0.085\textwidth}>{\centering\arraybackslash}m{0.085\textwidth}>{\centering\arraybackslash}m{0.085\textwidth}>{\centering\arraybackslash}m{0.085\textwidth}>{\centering\arraybackslash}m{0.085\textwidth}}
\toprule
 & \multicolumn{4}{c}{ Problem Solving }  & \multicolumn{4}{c}{ Commonsense Reasoning} & \multicolumn{2}{c}{ Reading Compreh.} & \\
\cmidrule(lr){2-5} \cmidrule(lr){6-9} \cmidrule(lr){10-11} 
 \multirow{1}{*}{\textbf{Selection Method}}& \textbf{ARC-E} & \textbf{ARC-C} & \textbf{MathQA} & \textbf{MMLU} & \textbf{O.B.QA} & \textbf{SIQA} & \textbf{W.G.} & \textbf{C.S.QA} & \textbf{BoolQ} & \textbf{RACE} & \textbf{Average} \\
\midrule
Random Sampling& 34.8 & 17.7 & 21.3 & 23.0 & 12.0 & 32.9 & 50.2 & 19.6 & 37.8 & 20.9 & 27.0 \\
Multi-actor collaboration (Ours) & \textbf{42.9} & \textbf{21.3} & \textbf{21.9} & \textbf{24.0} & \textbf{15.8} & \textbf{33.9} & \textbf{51.0} & \textbf{20.4} & \textbf{54.8} & \textbf{21.2} & \textbf{30.7} \\
\bottomrule
\end{tabular}
}
\end{table*}

\begin{table*}[tb]
\centering
\caption{This study compares the performance of training \textbf{8B model} from scratch on 25B tokens selected using random sampling versus multi-actor collaboration. Accuracy is reported for all tasks, with the highest value in each column shown in \textbf{bold}.}
\label{table:8B}
\vspace{1em}
\adjustbox{max width=1\textwidth}{
\begin{tabular}{l>{\centering\arraybackslash}m{0.085\textwidth}>{\centering\arraybackslash}m{0.085\textwidth}>{\centering\arraybackslash}m{0.085\textwidth}>{\centering\arraybackslash}m{0.085\textwidth}>{\centering\arraybackslash}m{0.085\textwidth}>{\centering\arraybackslash}m{0.085\textwidth}>{\centering\arraybackslash}m{0.085\textwidth}>{\centering\arraybackslash}m{0.085\textwidth}>{\centering\arraybackslash}m{0.085\textwidth}>{\centering\arraybackslash}m{0.085\textwidth}>{\centering\arraybackslash}m{0.085\textwidth}}
\toprule
 & \multicolumn{4}{c}{ Problem Solving }  & \multicolumn{4}{c}{ Commonsense Reasoning} & \multicolumn{2}{c}{ Reading Compreh.} & \\
\cmidrule(lr){2-5} \cmidrule(lr){6-9} \cmidrule(lr){10-11} 
 \multirow{1}{*}{\textbf{Selection Method}}& \textbf{ARC-E} & \textbf{ARC-C} & \textbf{MathQA} & \textbf{MMLU} & \textbf{O.B.QA} & \textbf{SIQA} & \textbf{W.G.} & \textbf{C.S.QA} & \textbf{BoolQ} & \textbf{RACE} & \textbf{Average} \\
\midrule
Random Sampling& 22.2 & 53.3 & 21.5 & 23.3 & 19.4 & 36.1 & 51.0 & 18.2 & 48.0 & 21.3 & 31.4 \\
Multi-actor collaboration (Ours) & \textbf{25.5} & \textbf{58.0} & \textbf{23.2} & \textbf{24.1} & \textbf{21.6} & \textbf{38.8} & \textbf{53.1} & \textbf{20.8} & \textbf{57.8} & \textbf{22.9} & \textbf{34.6} \\
\bottomrule
\end{tabular}
}
\end{table*}

\subsection{Generalization to 3.6B and 8B Models}
\label{app:generalization}
To evaluate the scalability of our approach, we conducted additional experiments training a 3.6 billion parameter model based on the LLaMA 3.2 architecture, which further demonstrates the scalability of our method. So far we have trained on 36 billion tokens and achieved strong performance, with plans to continue training with additional tokens according to scaling laws. As Table \ref{table:ablation_3B} shows, when compared to random selection, our method shows consistent performance improvements across all downstream tasks, achieving a 13.7\% increase in average accuracy—significantly higher than the 10.5\% improvement observed with the 1.3B models. Our results in Table \ref{table:8B} show that our methods outperform random selection approaches by 10.2\% for 8B LLaMA 3.2 architecture models.
Based on trends across three different model sizes (373M, 1.3B, 3.6B and 8B), our approach consistently outperforms random selection by over 10\% on average. This consistent advantage makes us believe that it suggests our method has strong potential for training even larger models, including those with 10B+ parameters.


\subsection{Details of Training Topic Classifier}
\label{sec:app:topic}

\begin{figure*}[tb]
    \centering
    \includegraphics[width=1\textwidth]{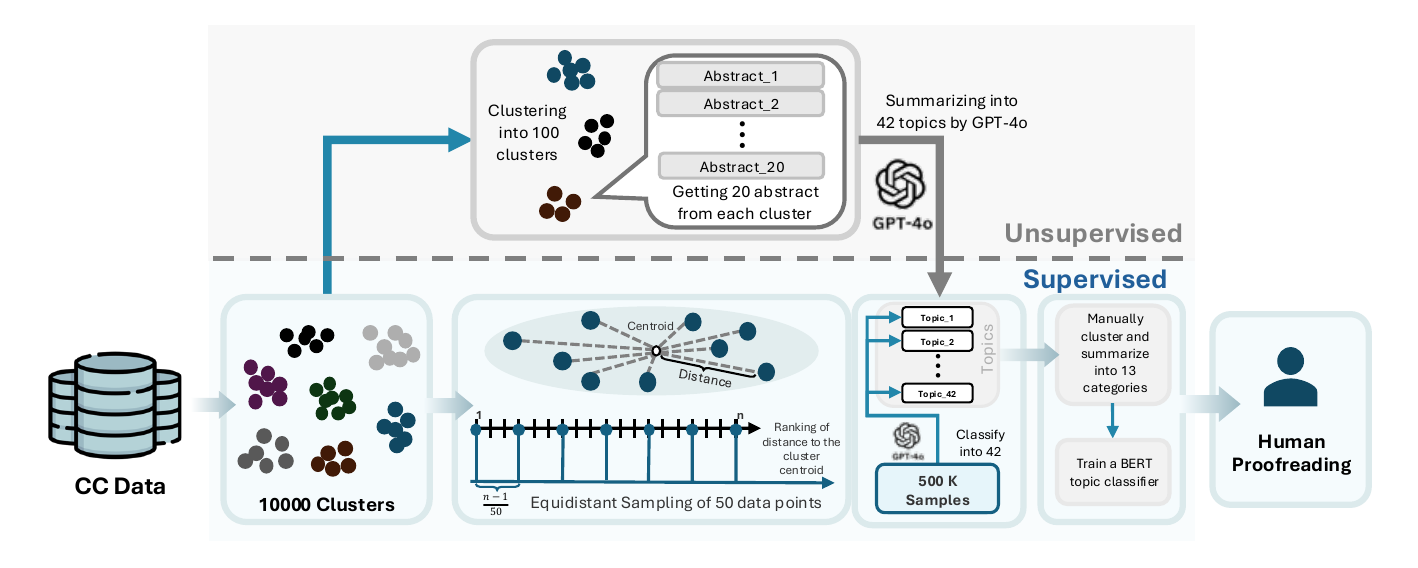}
    \caption{\textbf{Illustration of training process for topic classifier.} This diagram shows the process of training a BERT-based topic classifier using CommonCrawl data. 1.44 billion documents are clustered to generate topics. GPT-4o handles topic summarization and annotation, while a BERT model is trained to classify 13 topics, with humans doing final proofreading.}
    \label{fig:topic_classifier}
    \vspace{-0.1in}
\end{figure*}

As shown in Figure \ref{fig:topic_classifier}, we first cluster 1.4 billion documents obtained from Common Crawl~\citep{commoncrawl} into 10,000 clusters using KNN. And we use GPT-4o~\citep{gpt4o} to generate a summary for the content in each cluster. Additionally, we implement two parallel steps: unsupervised and supervised. In the unsupervised step, we perform secondary clustering of the 10,000 clusters into 100 clusters, from which we extract 20 summaries for each cluster. We utilize GPT-4o to extract category labels, refining these into a coherent hierarchical labeling system for the classification of 42 distinct topics.

In the supervised data processing step, leveraging Gopher cleaning ruls~\citep{gopher} and Min-Hash~\citep{broder1997resemblance} deduplication, we clean the whole datasets and cluster the datasets into 10,000 clusters. We then extract 50 equidistant samples from each cluster. This process yields approximately 500,000 data points, which we categorize into the aforementioned 42 topics by calling GPT-4o~\citep{gpt4o} using the prompt shown below:

\begin{figure*}[!htb]
    \centering
    \includegraphics[width=0.7\textwidth]{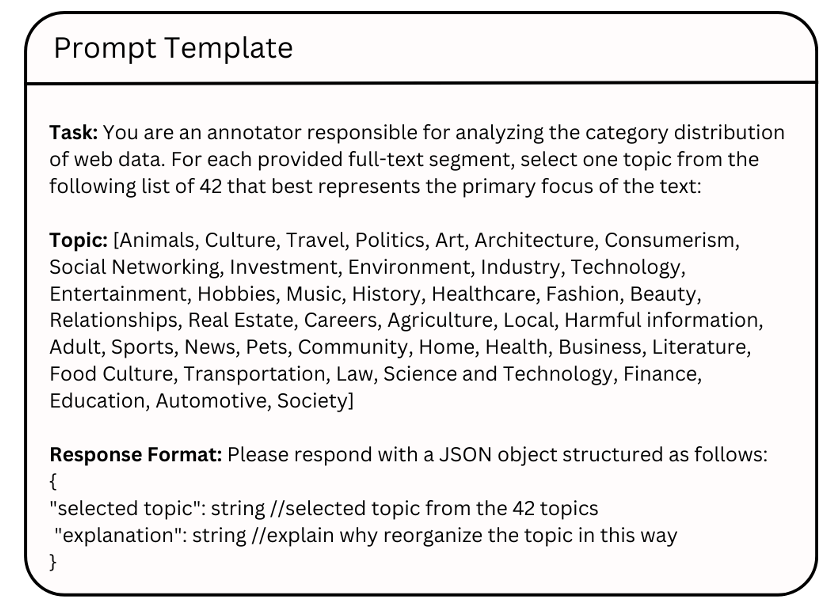}
    \caption{We illustrate the prompt construction process for GPT-4 to reorganize the topic of 500k data points.}
    \label{fig:my_image}
\end{figure*}

\begin{figure*}[!htb]
    \centering
    \includegraphics[width=1\textwidth]{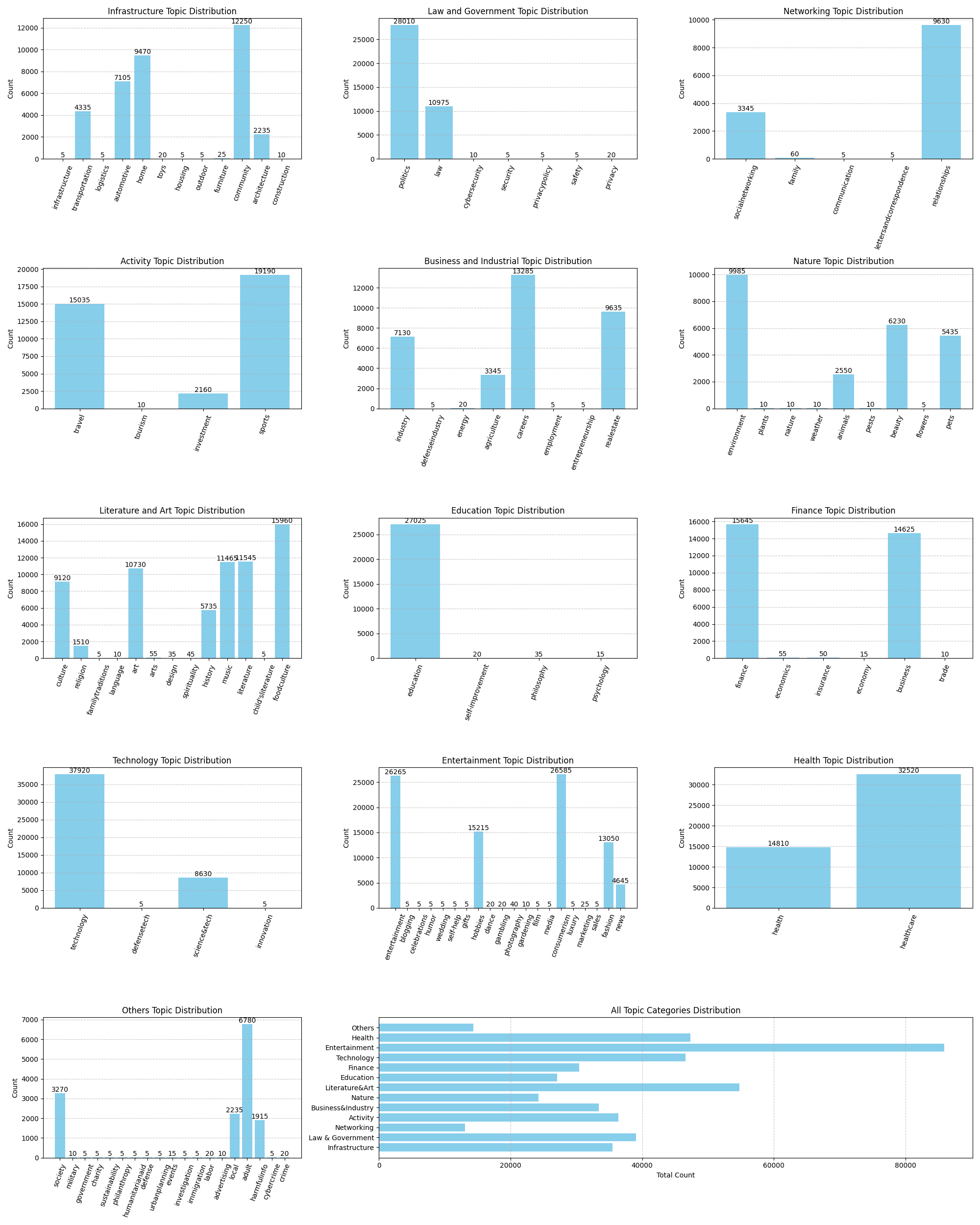}
    \caption{We illustrate the distribution of manually annotated and clustered data, which includes 13 topics: Infrastructure, Law and Government, Networking, Activity, Business and Industry, Nature, Literature and Art, Education, Finance, Technology, Entertainment, Health, and Others.}
    \label{fig:topic_distribution}
\end{figure*}

Since GPT-4o is not specialized for classification tasks, we obtain actual topic data with slightly more than 42 topics, as shown in Figure \ref{fig:topic_distribution}.
We then manually summarize the topics provided by GPT-4 into 13 categories, ensuring that the subtopics within each category shared similarities. The detailed category distributions appear in Figure \ref{fig:topic_distribution}, along with specific clustering information. Ultimately, we employ the annotated data to fine-tune a BERT-like regression model~\citep{bert}. Following model classification, we conduct human proofreading to ensure accuracy, and we present the final results below.

\subsection{Guidelines for Generalizing a New Criteria as an actor}

This section provides a detailed guidelines for incorporating a new criterion into our multi-actor system. The process is designed to ensure seamless integration and effective collaboration between existing and new actors.

\begin{algorithm*}[tb]
  \small
  
  \caption{Integrating a New Criterion into Multi-Actor Collaboration}
  \label{algo:new_actor}
  \begin{algorithmic}[1]
  \Require Sampled dataset $\mathcal{D}_\text{sample}$, pretraining dataset $\mathcal{D}_\text{train}$, reference dataset $D_\text{ref}$, existing Actors $\{\mathcal{A}_i\}_{i=1}^{3}$, scoring weights $\{\theta_i\}_{i=1}^{|\mathcal{D}|}$, memory $\mathcal{H}_{\mathcal{A}}$ for each Actor.  
    \State \textbf{Annotating Data for the New Criteria} 
    \State \hspace{1em} Sample data from the whole datasets, and annotate sampled dataset $\mathcal{D}_\text{sample}$ according to new criterion.
    \State \textbf{Training a Classifier for the New Criteria}
    \State \hspace{1em} Train a supervised classifier on $\mathcal{D}_\text{sample}$.
    \State \textbf{Defining the New Actor}
    \begin{itemize}
        \item \textbf{Action Space:} Sample and assess data points across subcategories of the new Criteria.
        \item \textbf{Memory:} Store prior scores and update based on model feedbacks.
    \end{itemize}
    \State \textbf{Labeling the Pretraining Dataset}
    \State \hspace{1em} Use trained classifier to label the whole training dataset $\mathcal{D}_\text{train}$ and store these labels in memory $\mathcal{H}_{\mathcal{A}_\text{new}}$.
  \State \textbf{Defining Actor Weights and Collaboration Strategy}
\begin{enumerate}
    \item Define the subcategory weights updating mechanism for $\mathcal{A}_\text{new}$ using Eq.~\ref{eq:agent_update}:
    \[
    w_{\text{new}}^j \gets (1-\eta_\text{new}) \cdot w_{\text{new}}^j + \eta_\text{new} \cdot R_\text{new}^j.
    \]
    \item Integrate $\mathcal{A}_\text{new}$ into the scoring function using Eq.~\ref{eq:coordinator}:
    \[
    S(x_i) = \theta_\text{Quality} S_\text{Quality}(x_i) + \theta_\text{Domain} S_\text{Domain}(x_i) + \theta_\text{Topic} S_\text{Topic}(x_i) + \theta_\text{new} S_\text{new}(x_i).
    \]
\end{enumerate}
  \State \textbf{Actor Initialization with Regression Techniques}
  \State \hspace{1em} Initialize the Actor weights $w_{\text{new}}$ using regression techniques (e.g., \texttt{RegMix}).
  \end{algorithmic}
\end{algorithm*}

Our framework offers several significant benefits when integrating a new actor. First, it offers \textbf{flexibility}, as the addition of new criteria can be performed independently of the core framework. This decoupling ensures that introducing new components does not require significant structural changes, allowing for smooth integration with minimal disruption to existing processes. Second, the approach is highly \textbf{scalable}. By enabling the training of new classifiers offline, the system can be easily adapted to handle a wide variety of data selection goals, which can evolve over time as new criteria emerge. Finally, the framework ensures \textbf{extensibility}, meaning it can seamlessly accommodate new objectives, whether simple or complex. This extensibility is key to maintaining efficient and effective collaboration across multiple actors, regardless of the size or complexity of the task at hand.

As demonstrated in Algorithm \ref{algo:multi_agent}, our framework seamlessly integrates a new actor through a series of straightforward steps. This approach ensures that the multi-actor framework remains flexible and effective in addressing diverse data selection objectives while preserving its collaborative efficiency.

\subsection{Connection of Multi-actor Collaborative Selection Method to Multi-agent RL}
\label{sec:app:marl}

The proposed multi-actor collaborative selection method is fundamentally inspired by the intelligent actor defined in \cite{russell2016artificial}, where the actor generally refers to an entity that perceives some status and map the observed status into actions. However, our framework also has many similarity compare with traditional multi-agent framework in reinforcement learning, where multiple actors work together to optimize a shared objective. In this section, we formally demonstrate the relationship between our framework and traditional multi-agent RL.

\subsubsection{Overall Definition in Reinforcement Learning Formulation}

We first clearly formulate each component of framework compare with components in general MARL framework for understanding the mechanism of our framework.
As our goal is to select the opt the \textbf{global action} at each step involves selecting a subset of data, \( \mathcal{D}_k \), from the entire dataset, \( \mathcal{D} \). This subset is used to update the model, where batches are drawn from \( \mathcal{D}_k \) for training. The \textbf{global state} is represented by the current model parameters, \( M \), which evolve as the model is trained. The \textbf{state transition} is formalized as $M' = \mathcal{O}(M, \mathcal{D}_k)$, 
where \( M' \) denotes the updated model after training on \( \mathcal{D}_k \).

The \textbf{reward function} measures the improvement in model performance on a reference task, \( \mathcal{T}_{\text{ref}} \), which serves as a proxy for the true downstream task \( \mathcal{T}_{\text{eval}} \). The reward is defined as:
\begin{equation*}
    R(M'| M, \mathcal{T}_{\text{ref}} ) = \mathbb{E}_{x \sim \mathcal{T}_{\text{ref}}}[-l(x; M')],
\end{equation*}
where $l(x; M')$ is the loss on $\mathcal{T}_{\text{ref}}$. For individual data points, the reward is estimated using influence functions:
\begin{equation*}
    r(x_i) = \text{Influence}(x_i, M, \mathcal{T}_{\text{ref}}).
\end{equation*}
This formulation links data selection directly to its impact on improving the model’s performance on $\mathcal{T}_{\text{ref}}$.

\subsubsection{Actor Design}

The estimation of value is as follows: The actor stores a reward estimation vector for each subset. The update rule is given by 
\begin{equation*}
    w_\mathcal{A}^j(t+1) = (1-\eta_\mathcal{A}) \cdot w_\mathcal{A}^j(t) + \eta_\mathcal{A} \cdot \bar{R}_\mathcal{A}^j.
\end{equation*}
The sliding average is used here because if all data in a subset were fully processed to compute $\bar{R}_\mathcal{A}^j$, there would be no need for a sliding average. However, since only a portion of the data is sampled, the estimate has higher variance, which is not favorable for training.
At the same time, the influence score itself is dynamic (even if the data remains constant, the model evolves). Averaging with outdated scores introduces bias. Therefore, the sliding average factor $\eta_\mathcal{A}$ strikes a 'bias-variance tradeoff'. We assume the score estimate for each data point $x_i$ in $\mathcal{D}_\mathcal{A}^j$ with respect to the dimension of interest for the actor, is given by $S_\mathcal{A}(x_i) = w_\mathcal{A}^j$ where $x_i \in \mathcal{D}_\mathcal{A}^j.$



\begin{algorithm*}[tb]
  \small
  \caption{Multi-actor collaborative data selection for LM pretraining}
  \label{algo:multi_agent}
  \begin{algorithmic}[1]
  \Require Training data $\mathcal{D}$, reference task $\mathcal{D}_\text{ref}$, main model $\mathcal{M}$, optimizer $\mathcal{O}$, total training steps $T$, selected size $k$,  update step $U$, Memory for each actor $\mathcal{H}_\mathcal{A}$
    \State Initialize model parameters for main model $\mathcal{M}$
    \State Initialize $\mathcal{D}_k$ as a size-k randomly sampled subset from $D$
    \For{$t = 1$ to $T$}
        \If{$t \bmod U = 0$}
            \For{each actor $\mathcal{A}$}
                \State Sample data points according to actor’s predefined sampling distribution
                \State Compute rewards $\mathcal{R}_\mathcal{M}(x_i;\mathcal{D}_\text{ref})$ for each sampled data point $x_i$
                \State \textbf{Update actor weight} $\mathbf{w_{\mathcal{A}}} \leftarrow \mathbf{w_{\mathcal{A}}} + \eta_{\mathcal{A}} \cdot \mathbf{\overline{R}_{\mathcal{A}}}$
            \EndFor
            \State Compute actor score $\overline{R}_{\mathcal{A}}$ and average score $\overline{R}$ according to Eq. \ref{eq:collaborative_score}
            \State \textbf{Update collaborative weight} $\theta_\mathcal{A} \leftarrow \theta_\mathcal{A} + \eta_\mathcal{A} \cdot (\overline{R}_{\mathcal{A}} - \overline{R}).$ 
            \State Calculate coordinator score $S(x_i)$ for $x_i \in \mathcal{D}$ according to Eq. \ref{eq:coordinator}
            \State Select dataset for next training stage $\mathcal{D}_k \leftarrow \text{Top-}k(S(x_i))$ for $x_i \in \mathcal{D}$
        \EndIf
        \State Sample a batch of data $B$ from $\mathcal{D}_k$
        \State \textbf{Update Main Model} $\mathcal{M} \gets \mathcal{O}(\mathcal{M}, B)$
    \EndFor
\end{algorithmic}
\end{algorithm*}

\subsubsection{Multi-actor Collaboration}

Assume that the score of a single data point in the reference task is obtained as a weighted sum of multiple components. The total score for each data point is given by Equation~\ref{eq:collaborative_score} as:
\begin{equation*}
    S(x_i) = \sum_{\mathcal{A} } \theta_\mathcal{A} \cdot S_\mathcal{A}(x_i),
\end{equation*}
where $\theta_\mathcal{A}$ are collaborative weights. A central coordinator adjusts these weights over time based on the actors’ contributions to the overall reward:
\begin{equation*}
    \theta_\mathcal{A}(t+1) = \theta_\mathcal{A}(t) + \eta_\mathcal{A} (\bar{R}_\mathcal{A} - \bar{R}),
\end{equation*}
where \( \bar{R}_\mathcal{A} \) is the actor’s average reward, and \( \bar{R} \) is the global average reward:
\begin{equation*}
    \bar{R}_\mathcal{A} = \frac{1}{n} \sum_{j=1}^n w_\mathcal{A}^j \cdot \bar{R}_\mathcal{A}^j, \quad \bar{R} = \frac{1}{3} \sum_{\mathcal{A} } \bar{R}_\mathcal{A}.
\end{equation*}
We consider three possible cases for our framework, comparing its relationship with traditional optimization problem.
\begin{itemize}[topsep=5pt, leftmargin=*]
    \item \textbf{Single-actor case}: If only one actor is involved, $\theta$ becomes irrelevant, reducing the problem to a classical optimization scenario where the actor greedily selects the optimal data based on one criteria.
    \item \textbf{Multi-actor competitive mechanism}: When multiple actors are present, $\theta$ reflects each actor’s capability. Selecting the best-performing actor for decision-making introduces a heuristic competitive mechanism, building upon the classical optimization framework.
    \item \textbf{Multi-actor collaborative mechanism}: Alternatively, when multiple actors are involved, $\theta$ can be used to weigh each actor's contributions for decision-making. This introduces a smoother heuristic cooperative mechanism, extending the classical optimization framework by leveraging weighted collaboration. This heuristic cooperative mechanism dynamically adjusts the influence of each actor based on the model's current preferences, enabling more effective data filtering decisions.
\end{itemize}
In practice, we choose to use the multi-actor collaborative mechanism for data selection. We have added comparisons with single-actor and competitive mechanisms in Table~\ref{table:dynamic} to further elaborate the effectiveness of collaboration.

\begin{table*}[tb]
\centering
\caption{This ablation study examines the performance of various combinations of actor collaboration (\textbf{Actor}) and dynamic collaborative weight update (\textbf{Dynamic}). Accuracy is reported for all tasks, with the highest value in each column shown in \textbf{bold}.}
\label{table:dynamic}
\vspace{1em}
\adjustbox{max width=1\textwidth}{
\begin{tabular}{l>{\centering\arraybackslash}m{0.085\textwidth}>{\centering\arraybackslash}m{0.085\textwidth}>{\centering\arraybackslash}m{0.085\textwidth}>{\centering\arraybackslash}m{0.085\textwidth}>{\centering\arraybackslash}m{0.085\textwidth}>{\centering\arraybackslash}m{0.085\textwidth}>{\centering\arraybackslash}m{0.085\textwidth}>{\centering\arraybackslash}m{0.085\textwidth}>{\centering\arraybackslash}m{0.085\textwidth}>{\centering\arraybackslash}m{0.085\textwidth}>{\centering\arraybackslash}m{0.085\textwidth}>{\centering\arraybackslash}m{0.085\textwidth}}
\toprule
 & & \multicolumn{4}{c}{ Problem Solving }  & \multicolumn{4}{c}{ Commonsense Reasoning} & \multicolumn{2}{c}{ Reading Compreh.} & \\
\cmidrule(lr){3-6} \cmidrule(lr){7-10} \cmidrule(lr){11-12} 
 {\textbf{actor}} & {\textbf{Dynamic }} & \textbf{ARC-E} & \textbf{ARC-C} & \textbf{MathQA} & \textbf{MMLU} & \textbf{O.B.QA} & \textbf{SIQA} & \textbf{W.G.} & \textbf{C.S.QA} & \textbf{BoolQ} & \textbf{RACE} & \textbf{Average} \\
\midrule
with & with & \textbf{65.8} & \textbf{31.5} & \textbf{23.0} & \textbf{26.6} & \textbf{24.6} & \textbf{39.9} & \textbf{54.1} & \textbf{20.1} & \textbf{60.4} & \textbf{30.5} & \textbf{37.7} \\
with & without & 59.4 & 26.3 & 21.3 & 25.1 & 20.5 & 38.9 & 52.9 & 19.8 & 58.1 & 28.3 & 35.1 \\
without & - & 59.2 & 26.1 & 20.3 & 25.4 & 21.3 & 39.1 & 52.6 & \textbf{20.1} & 56.5 & 29.1 & 35.0 \\
\bottomrule
\end{tabular}
}
\end{table*}

\subsection{Detailed Analysis of Computational Overhead} 

In this subsection, we compare the computational overhead of our multi-actor collaboration framework with baseline approaches. The analysis focuses on three aspects: \textbf{offline computation efficiency}, \textbf{online computation efficiency}, and \textbf{overall FLOPs requirements}. Table \ref{table:computational_overhead} summarizes these comparisons.

\begin{table*}[ht]
\caption{Comparison of Computational Overhead Across Methods}
\centering
\begin{tabular}{lccc}
\toprule
\textbf{Selection Method} & \textbf{Offline Computation} & \textbf{Online Computation} & \textbf{Overall} \\
 & \textbf{Cost (FLOPs)} & \textbf{Cost  (FLOPs)} & \textbf{ (FLOPs)}\\
\midrule
Qu-Rating~\citep{qurating} & $7.13 \times 10^{20}$ & N.A. & $7.13 \times 10^{20}$ \\
MATES~\citep{mates} & N.A. & $1.99 \times 10^{20}$ & $1.99 \times 10^{20}$\\
Multi-actor collaboration (ours) & $9.91 \times 10^{19}$ & $1.19\times 10^{18}$ & $1.00 \times 10^{20}$ \\
\bottomrule
\end{tabular}
\label{table:computational_overhead}
\end{table*}

\subsubsection{Offline Computation Efficiency}

Our method achieves superior offline efficiency by requiring only $9.91 \times 10^{19}$ FLOPs for a one-time dataset labeling process using a 109M BERT-based model for inference. This is nearly an order of magnitude more efficient than Qu-Rating, which consumes $7.13 \times 10^{20}$ FLOPs due to its reliance on a larger 1.3B Sheared-LLaMA model. MATES does not utilize offline computation, relying solely on online updates, which avoids this cost but limits its flexibility and scalability. The offline labeling in our method ensures robust initial scores for large-scale datasets while laying the groundwork for efficient online updates.

\subsubsection{Online Computation Efficiency}

For adaptive online updates, both our approach and MATES compute influence scores with $1.19\times 10^{18}$ FLOPs. However, MATES involves labeling the entire dataset with a 109M BERT-based model in every round, amounting to $1.98 \times 10^{20}$ FLOPs across four data selection stages. In contrast, our method avoids re-labeling the entire dataset, significantly reducing the computational cost by focusing on labeling the large pretraining datasets only once.


Overall, our approach cuts the computational cost in half compared to MATES and requires only about 1/7 of the computational resources used by Qu-Rating.







\subsection{Analysis of Ablation Study}
\subsubsection{Analysis of actor Roles on Different Type of Tasks}
\label{app:ablation:actor}
\begin{table*}[tb]
\centering
\caption{This ablation study examines the performance of various combinations of actor collaboration and update mechanisms. All models are in 373M LLaMA2 architecture. Accuracy is reported for all tasks, with the highest value in each column shown in \textbf{bold}. 
}\label{table:ablation}
\vspace{1em}
\adjustbox{max width=1\textwidth}{
\begin{tabular}{l>{\centering\arraybackslash}m{0.085\textwidth}>{\centering\arraybackslash}m{0.085\textwidth}>{\centering\arraybackslash}m{0.085\textwidth}>{\centering\arraybackslash}m{0.085\textwidth}>{\centering\arraybackslash}m{0.085\textwidth}>{\centering\arraybackslash}m{0.085\textwidth}>{\centering\arraybackslash}m{0.085\textwidth}>{\centering\arraybackslash}m{0.085\textwidth}>{\centering\arraybackslash}m{0.085\textwidth}>{\centering\arraybackslash}m{0.085\textwidth}>{\centering\arraybackslash}m{0.085\textwidth}}
\toprule
 & \multicolumn{4}{c}{ Problem Solving }  & \multicolumn{4}{c}{ Commonsense Reasoning} & \multicolumn{2}{c}{ Reading Compreh.} & \\
\cmidrule(lr){2-5} \cmidrule(lr){6-9} \cmidrule(lr){10-11} 
 \multirow{1}{*}{\textbf{Selection Method}}& \textbf{ARC-E} & \textbf{ARC-C} & \textbf{MathQA} & \textbf{MMLU} & \textbf{O.B.QA} & \textbf{SIQA} & \textbf{W.G.} & \textbf{C.S.QA} & \textbf{BoolQ} & \textbf{RACE} & \textbf{Average} \\
\midrule
Quality\&Domain\&Topic actor & \textbf{57.9} & \textbf{24.7} & \textbf{21.9} & 25.4 & \textbf{20.2} & \textbf{37.9} & \textbf{52.6} & \textbf{20.4} & 59.6 & \textbf{29.4} & \textbf{35.0} \\
without collaboration update & 47.9 & 20.4 & 21.0 & 25.1  & 17.2  & 37.3 &  51.3 & 20.0 & 56.5 & 28.3  & 32.5 \\
\midrule
Domain\&Quality actor & 55.1 & 18.6 & 21.7 & 24.4 & 17.4 & 37.1 & 51.2 & 19.8 & \textbf{61.7}& 28.2 & 33.5 \\
Topic\&Quality actor & 56.2 & 24.4 & 21.8 & 25.2 & 19.4 & 36.3 & 49.0 & 19.7 & 56.1 & 28.5 & 33.6 \\
Domain\&Topic actor & 44.6 & 18.3 & 21.7 & 25.7 & 16.2 & 36.6 & 51.9 & 19.9 & 61.6 & 27.8 & 32.4 \\
\midrule
Quality actor & 53.0 & 24.7 & 21.8 & 25.5 & 18.0 & 36.3 & 49.5 & 18.1 & 57.0 & 28.0 & 32.9 \\
Domain actor & 44.1 & 19.1 & 20.8 & 25.6 & 16.6 & 36.8 & 52.0 & 19.7 & 56.7 & 28.2 & 32.0 \\
Topic actor & 42.7 & 19.2 & 21.0 & \textbf{27.0} & 17.4 & 37.1 & 50.7 & 19.7 & 54.6 & 28.5 & 31.8 \\
\midrule
No actor & 42.5 & 20.0 &21.1 & 23.8 & 14.6	& 35.9 & 50.1 &	18.8 & 56.1	& 27.9 & 31.1
 \\
\bottomrule
\end{tabular}
}
\end{table*}

We show the actor ablation study conducted on 373M LLaMA2 models Table ~\ref{table:ablation} as well as 1.3B LLaMA2 models Table~\ref{table:ablation_1.3B}.

\underline{First}, the ablation study underscores \textit{the importance of each actor in achieving optimal performance} across the training tasks. When the quality, domain, and topic actors are used together, the model performs best, highlighting the benefits of their combined use, as shown in Table \ref{table:ablation_1.3B}. In terms of evaluating \textit{the performance of each actor}, we find that the quality actor outperforms other single-actor configurations, excelling in problem solving tasks. However, its performance drops on tasks requiring domain knowledge or contextual understanding. Here, the domain and topic actors play a crucial role, as they excel in these areas. Despite this, neither performs well on problem solving tasks, except for the topic actor, which significantly improves MMLU performance, indicating that topic diversity may benefit such tasks. In terms of evaluating \textit{the combination of the actors}, we find that removing any actor noticeably reduces overall accuracy, though the impact varies. Excluding the quality actor leads to the largest drop, significantly affecting performance in problem solving tasks, and commonsense reasoning tasks like OpenbookQA. This highlights the quality actor’s vital role in reasoning and problem-solving. Similarly, excluding the topic actor causes a performance drop in ARC-Challenge and a significant reduction in MMLU, emphasizing its importance in tasks covering diverse subjects; removing the domain actor results in a performance drop in commonsense reasoning tasks, underscoring its key contribution to these areas.
\underline{Second}, the ablation study verifies the design and implementation of \textit{the collaborative dynamic adjustment of the actors' weights} (introduced \S \ref{sec:methods:collaborate}) for efficient data selection. Concretely, in this variant, all actors were initialized with equal weights, which remained fixed throughout training without adjusting for individual actor performance. Surprisingly, this fixed-weight approach (equal to random sampling) resulted in a significant drop in overall performance compared to the dynamic weighting used in the collaborative update framework, as shown in Table~\ref{table:ablation_1.3B}. We believe this result from the ablation study is a strong indicator that the dynamic adjustment of the celebration mechanism is essential for efficient data selection. 

\subsubsection{Ablation Study on Reference Tasks Selection}
\begin{table*}[tb]
\centering

\caption{This ablation study examines the performance of various combinations of reference task. Accuracy is reported for all tasks, with the highest value in each column shown in \textbf{bold}. 
}\label{table:ablation_reference}
\vspace{1em}
\adjustbox{max width=1\textwidth}{
\begin{tabular}{l>{\centering\arraybackslash}m{0.085\textwidth}>{\centering\arraybackslash}m{0.085\textwidth}>{\centering\arraybackslash}m{0.085\textwidth}>{\centering\arraybackslash}m{0.085\textwidth}>{\centering\arraybackslash}m{0.085\textwidth}>{\centering\arraybackslash}m{0.085\textwidth}>{\centering\arraybackslash}m{0.085\textwidth}>{\centering\arraybackslash}m{0.085\textwidth}>{\centering\arraybackslash}m{0.085\textwidth}>{\centering\arraybackslash}m{0.085\textwidth}>{\centering\arraybackslash}m{0.085\textwidth}}
\toprule
 & \multicolumn{4}{c}{ Problem Solving }  & \multicolumn{4}{c}{ Commonsense Reasoning} & \multicolumn{2}{c}{ Reading Compreh.} & \\
\cmidrule(lr){2-5} \cmidrule(lr){6-9} \cmidrule(lr){10-11} 
 \multirow{1}{*}{\textbf{Reference Tasks}}& \textbf{ARC-E} & \textbf{ARC-C} & \textbf{MathQA} & \textbf{MMLU} & \textbf{O.B.QA} & \textbf{SIQA} & \textbf{W.G.} & \textbf{C.S.QA} & \textbf{BoolQ} & \textbf{RACE} & \textbf{Average} \\
\midrule
LAMBADA\&SQuAD\&Jeopardy & \textbf{65.8} & \textbf{31.5} & 23.0 & 26.6 & 24.6 & 39.9 & 54.1 & 20.1 & \textbf{60.4} & \textbf{30.5} & \textbf{37.7} \\
LAMBADA & 64.3 & 31.2 & 22.3 & \textbf{26.8} & 23.5 & 39.6 & \textbf{54.6} & 20.4 & 59.6 & 30.1 & 37.2 \\
SQuAD & 65.1 & 30.9 & \textbf{23.4} & 25.9 & \textbf{24.9} & 40.1 & 53.8 & 21.2 & 59.1 & 29.3 & 37.4 \\
Jeopardy & 63.9 & 30.3 & 23.6 & 26.3 & 24.1 & \textbf{40.7} & 54.5 & \textbf{21.8} & 59.1 & 30.2 & 37.5 \\
Random selection & 54.6 & 23.0 & 22.1 & 24.9 & 18.8 & 40.3 & 52.9 & 21.5 & 53.0 & 29.8 & 34.1 \\
\bottomrule
\end{tabular}
}
\end{table*}

In our experiments of selecting reference tasks in Table \ref{table:ablation_reference}, we observe that while the choice of reference tasks can influence performance, the impact on average performance is marginal (within 0.5 points). Using different reference tasks consistently leads to a significant improvement in average performance compared to random data selection, demonstrating that our method is not sensitive to the choice of reference tasks.



\subsection{Details of Pretraining}
\label{sec:app:pretrain}
\subsubsection{Details of Pretraining Models Architecture}
The specific architecture of pretraining model is shown in Table \ref{tab:model_architecture}.
Each model was trained on 32x NVIDIA A800, employing a global batch size of $4 \times 2^{20}$ tokens and completing 7,500 steps in about 14 hours. 
The average token processing rate per GPU was about 20,000 tokens per second. 
The learning rate was set to $5 \times 10^{-5}$, and the Adam optimizer was employed with hyperparameters ($\beta_1=0.9, \beta_2=0.95, \epsilon=10^{-8}$).

\begin{table*}[!htb]
\scriptsize 
\centering
\caption{Architecture of pre-trained decoder-only models.}
\label{tab:model_architecture}
\vspace{0.5em} 
\begin{tabular}{@{}lllll@{}}
\toprule
Hyperparameter & 370M Model Value & 1.3B Model Value & 3.6B Model Value & 8B Model     \\ \midrule
Vocabulary Size            & 32,000           & 32,000          & {128,256}         & {128,256}         \\
MLP Ratio                  & 8/3              & 8/3             & {8/3}             & {3.5}             \\
Hidden Dimension Size      & 2048             & 1024            & {3072}            & {4096}            \\
Number of Layers           & 24               & 24              & {28}              & {32}              \\
Number of Attention Heads  & 16               & 8               & {24}              & {32}              \\
Number of KV Attention Heads & 16            & 8               & {8}               & {8}               \\
RoPE Base                  & 10,000           & 10,000          & {500,000}         & {500,000}         \\
Maximum Context Window Length & 1024         & 1024            & {1024}            & {1024}            \\
Number of Parameters       & 373,867,520 (370M) & 1,345,423,360 (1.3B) & {3,606,752,256 (3.6B)} & {8,030,261,248 (8B)} \\ \bottomrule
\end{tabular}
\end{table*}

\subsubsection{Details of Baseline Method Implementation}
\label{sec:app:baseline}

Regarding the classifier methods, QuRating~\citep{qurating} and DSIR~\citep{dsir}, we implement QuRating by downloading the open-source checkpoint from Hugging Face. Notably, the released model has a context length of 4096, whereas ours is 1024. However, this discrepancy does not impact our testing tasks, as our maximum of 5-shot examples remains within the 1024 limit. Despite this, we have totally similar model configuration as well as the total number of training tokens with all the checkpoints we downloaded. Similarly, the replication of PPL is based on the publicly available checkpoint from the original paper. For FineWeb-Edu~\citep{lozhkov2024fineweb-edu}, we download the quality scorer to label all the training data from SlimPajama datasets, and adopt the methodology described in the corresponding publication and train all the model from scratch.

Domain mixing refers to the technique of combining data from different sources or domains to enhance the diversity and robustness of a model's training dataset. In our implementation, we apply various mixing methods: DoReMi~\citep{doremi}, DOGE~\citep{doge}, DMLaw~\citep{dmlaw}, and RegMix~\citep{regmix}. Each method contributes distinct proportions of data from specific domains, as reflected in the domain weights presented in Table \ref{tab:domain_weights}. Notably, the weights indicate the percentage of contributions from each domain.

\begin{table*}[htb]
\centering
\caption{Exact domain weights (\%) on SlimPajama obtained by data mixing methods.  Abbreviations: C.C. = CommonCrawl, Wiki = Wikipedia, StackEx. = StackExchange
}\label{tab:domain_weights}
\vspace{1em}
\adjustbox{max width=1\textwidth}{
\footnotesize 
\begin{tabular}{l>{\centering\arraybackslash}m{0.06\textwidth}>{\centering\arraybackslash}m{0.06\textwidth}>{\centering\arraybackslash}m{0.06\textwidth}>{\centering\arraybackslash}m{0.06\textwidth}>{\centering\arraybackslash}m{0.06\textwidth}>{\centering\arraybackslash}m{0.06\textwidth}>{\centering\arraybackslash}m{0.06\textwidth}}
\toprule
{ \textbf{Mixing Method}} & \textbf{C.C.} & \textbf{C4} & \textbf{GitHub} & \textbf{Books} & \textbf{ArXiv} & \textbf{Wiki} & \textbf{StackEx.} \\ \midrule
{ SlimPajama} & 52.20 & 26.70 & 5.20 & 4.20 & 4.60 & 3.80 & 3.30 \\ 
{ DoReMi}    & 38.11 & 11.41 & 6.54 & 8.19 & 4.24 & 23.07 & 8.47 \\ 
{ DOGE}      & 21.35 & 26.93 & 7.03 & 4.50 & 8.80 & 14.82 & 16.58 \\ 
{ DMLaw}     & 12.50 & 25.00 & 14.06 & 9.38 & 25.00 & 1.56 & 12.50 \\ 
{ RegMix}    & 17.37 & 51.03 & 0.23 & 0.23 & 0.08 & 29.77 & 1.27 \\ 
\bottomrule
\end{tabular}
}
\end{table*}

For the reproduction of MATES~\citep{mates}, we start by utilizing Random-Slimpajama at the 1500th training step as our primary pretraining model and fine-tune the BERT-base from the original thesis as our data influence model. During the training of the data influence model, we uniformly sample 1/13 of the data as hold-out data from each area of our dataset and employ LAMBADA~\citep{paperno2016lambada} as a reference task, following the MATES methodology. Ultimately, we use the trained BERT-base data influence model to predict the entire training dataset, selecting the top 1/20 as our pretraining data. This selection process is executed using the Gumbel-Top-k algorithm~\citep{gumbel_top}, consistent with MATES. We leverage a four-step updates similar to the original paper, and conduct the above implementation at 1500th, 3000th, 4500th and 6000th model training steps using the current models.

\subsubsection{Details actor Weight Initialization}
\label{sec:app:weight_initialize}

For the \textit{domain actor}, we use the document's meta-information, label the data with domain information and save this into the domain actor's memory, where the domain $\text{Domain}(x_i)$ belongs to one of ArXiv, Book, Wikipedia, CommonCrawl, GitHub, StackExchange, C4. 

For \textit{quality actor}, we adopt the FineWeb-Edu quality scorer~\citep{lozhkov2024fineweb-edu}, which is fine-tuned as a BERT-like regression model~\cite{merrick2024arctic} using Llama3-70B-Instruct annotated 500k examples. This will give out a successive quality score $\text{Quality}(x_i) \in \mathcal{R}^{[0,5]}$ with higher score represent higher quality. We then map this score into five quality intervals $\{I_j\}_{j=1}^5$, as
\begin{equation*}
\small
\begin{split}
\text{Quality}(x_i) \mapsto I_j = 
\begin{cases}
[j-1, j), & \text{if } \text{Quality}(x_i) \in [j-1, j), \\
         & \quad j = 1, 2, 3, 4 \\
[4, 5], & \text{if } \text{Quality}(x_i) \in [4, 5], \\
        & \quad j = 5
\end{cases}
\end{split}
\end{equation*}

We store the quality interval corresponding to each data point in the quality actor’s memory. 

For the \textit{topic actor}, due to the absence of a suitable pretrained model for topic classification and labeling, we designed a classification schema using 1.44 billion documents collected by the Common Crawl project~\citep{commoncrawl} and fine-tuned a BERT-like regression model on 500k GPT-4o annotated samples, the overall pipeline is depicted in Figure \ref{fig:topic_classifier}. Further details on the topic classification approach and BERT model training are provided in \S \ref{sec:app:topic}. Using this topic classifier, we categorize each document into one of 13 topics: Activity, Education, Entertainment, Finance, Health, Business and Industrial, Infrastructure, Literature and Art, Nature, Others, Law and Government, Networking, Technology, and store the topic information in the topic actor’s memory.

\begin{figure*}[ht]
    \centering
    \subfigure[Topic]{\includegraphics[width=.60\textwidth]{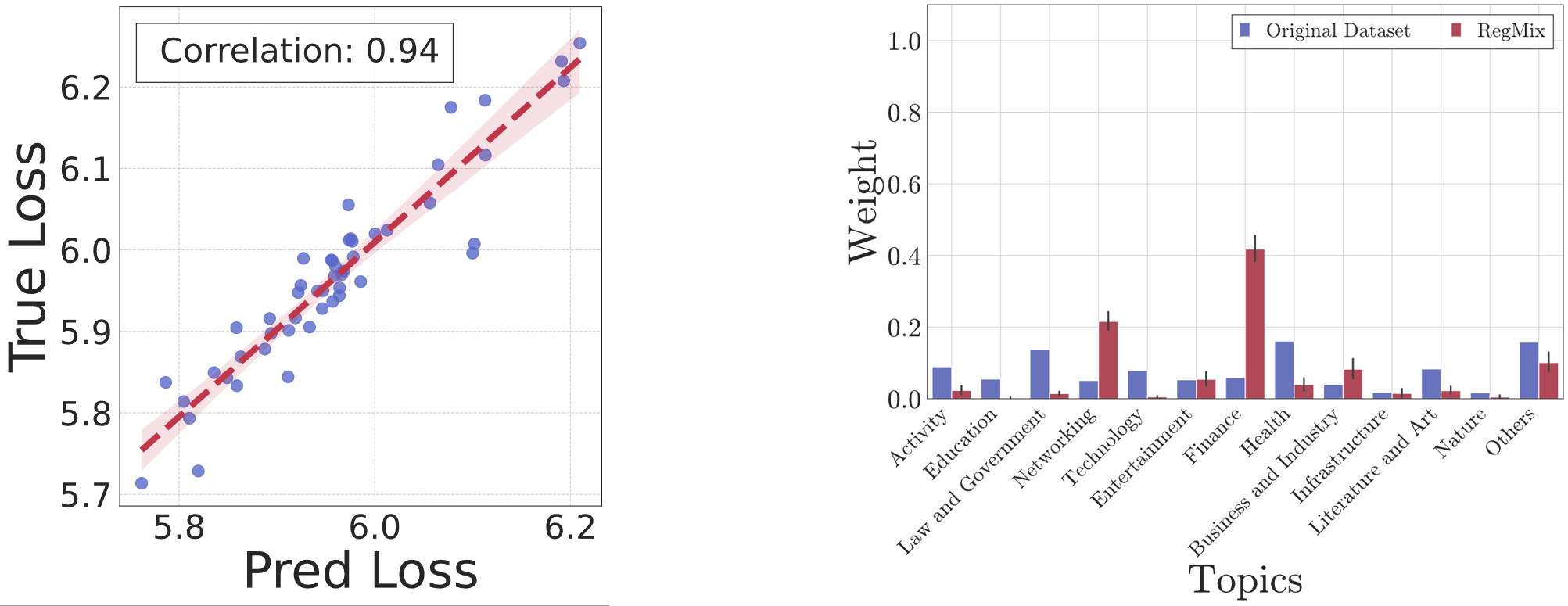}} \\
    \subfigure[Quality Interval]{\includegraphics[width=.60\textwidth]{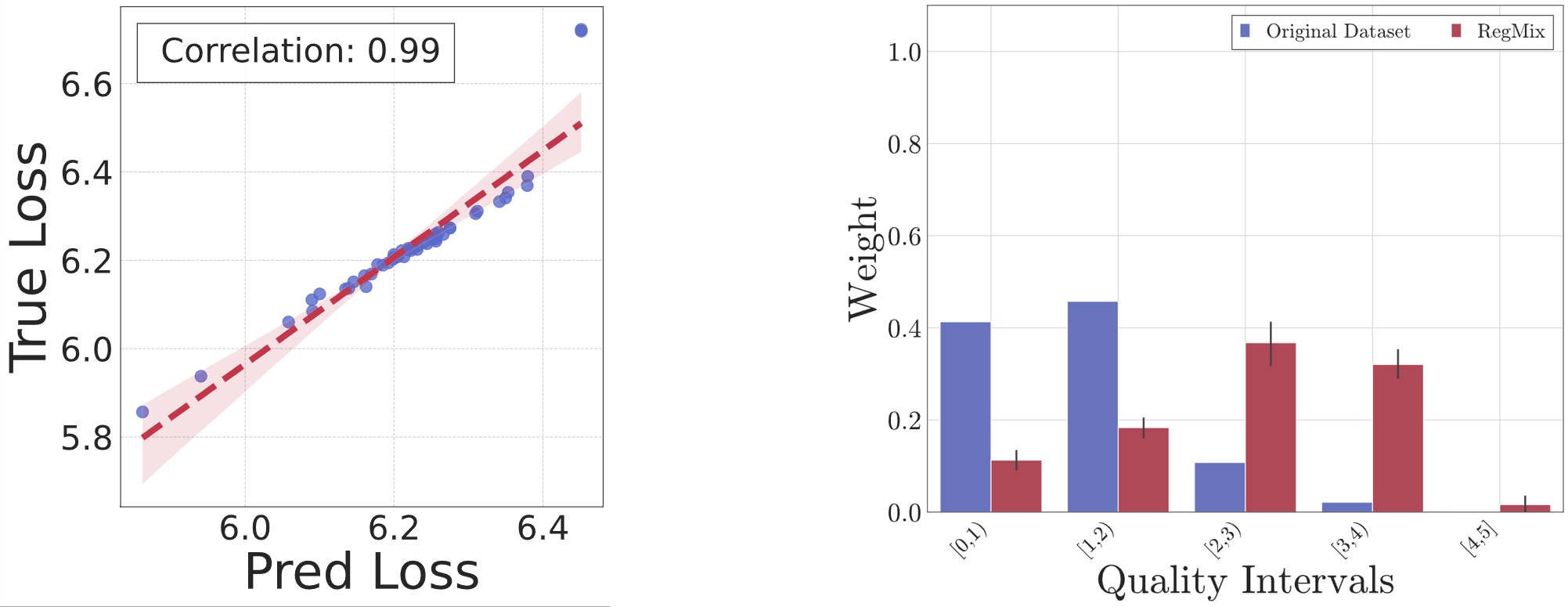}} \\
    \subfigure[Domain]{\includegraphics[width=.60\textwidth]{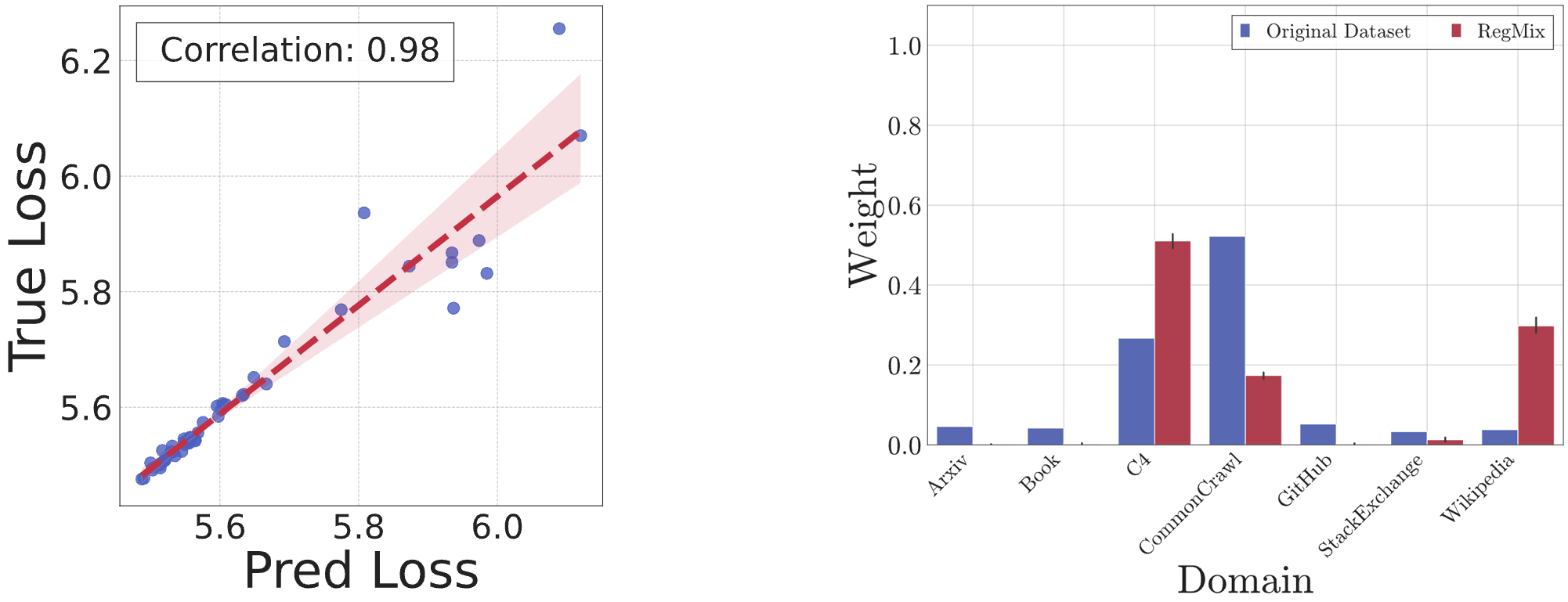}}   
    \caption{We present the results of the LightGBM regression analysis and Spearman correlation regarding the loss prediction performance and the weights of each candidate data-(a) Topic, (b) Quality Interval and (c) Domain after mixture across all categories.}
    \label{fig:lgcn}
\end{figure*}

We employ an actor weight initialization technique within the RegMix~\citep{regmix} framework, which is crucial for the effective training of proxy models. Our dataset is organized into three distinct categories: domain, quality, and topic. For each category, we initialize the data weights based on the original proportions across 512 configurations and subsequently train a TinyLlama-1M with 1 billion tokens as a proxy model for each configuration. We evaluate this model on previously unseen data mixtures, specifically using validation set loss, following RegMix, for assessment. We then fit a regression model based on the performance results of the 512 proxy models to predict the optimal data mixture for training large-scale LMs. The results of the LightGBM regression analysis and Spearman correlation of the loss prediction performance are presented in \ref{fig:lgcn}.

Upon training the regression model, we systematically investigate the entire spectrum of potential data mixtures by utilizing the trained model to predict the target values for each candidate mixture. This process allows us to identify the input that produces the optimal target value. Following the simulation and identification of the most effective data mixture, we then generalize this top-ranked configuration for large-scale model training, incorporating a significantly larger volume of tokens.


\subsection{Full Experimental Results}
\label{sec:app:full_results}
\begin{table*}[!htb]
\centering
\caption{Table Showing Various Selection Methods and Their Scores with Changes. We report accuracy for all tasks, and \textbf{bold} the best result in each column. Abbreviations: O.B.QA = OpenbookQA W.G. = WinoGrande, C.S.QA = CommonSenseQA, Compreh. = Comprehensions
}\label{table:zershot}
\vspace{1em}
\adjustbox{max width=0.9\textwidth}{
\begin{tabular}{l>{\centering\arraybackslash}m{0.085\textwidth}>{\centering\arraybackslash}m{0.085\textwidth}>{\centering\arraybackslash}m{0.085\textwidth}>{\centering\arraybackslash}m{0.085\textwidth}>{\centering\arraybackslash}m{0.085\textwidth}>{\centering\arraybackslash}m{0.085\textwidth}>{\centering\arraybackslash}m{0.085\textwidth}>{\centering\arraybackslash}m{0.085\textwidth}>{\centering\arraybackslash}m{0.085\textwidth}>{\centering\arraybackslash}m{0.085\textwidth}>{\centering\arraybackslash}m{0.085\textwidth}}
\toprule
 & \multicolumn{4}{c}{ Problem Solving }  & \multicolumn{4}{c}{ Commonsense Reasoning} & \multicolumn{2}{c}{ Reading Compreh.} & \\
\cmidrule(lr){2-5} \cmidrule(lr){6-9} \cmidrule(lr){10-11} 
 \multirow{1}{*}{\textbf{Selection Method}}& \textbf{ARC-E} & \textbf{ARC-C} & \textbf{MathQA} & \textbf{MMLU} & \textbf{O.B.QA} & \textbf{SIQA} & \textbf{W.G.} & \textbf{C.S.QA} & \textbf{BoolQ} & \textbf{RACE} & \textbf{Average} \\
\midrule
\textbf{Random sample} & & & & & & & & & & & \\
Uniform-30B & 54.3 & 23.4 & 22.3 & 23.9 & 18.6 & 39.8 & 52.8 & 19.2 & 55.4 & 30.0 & 34.0 \\
Uniform-60B & 55.2 & 24.6 & 22.5 & 23.4 & 21.0 & 39.7 & 51.9 & 19.5 & 59.8 & \textbf{33.1} & 35.1 \\
\midrule
\textbf{Perplexity-based data selection} & & & & & & & & & & & \\
PPL & 49.3 & 20.1 & 22.4 & 23.6 & 16.2 & 36.0 & 48.1 & 18.8 & 61.4 & 29.3 & 32.5 \\
\midrule
\textbf{Classifier-based data selection} & & & & & & & & & & & \\
QuRating-Facts & 56.1 & 23.3 & 22.4 & 24.8 & 21.6 & 39.2 & 54.1 & 19.9 & 61.5 & 31.6 & 35.5 \\
QuRating-Req & 54.9 & 24.4 & 23.2 & 25.2 & 21.4 & 38.1 & 54.5 & 20.6 & \textbf{61.6} & 31.3 & 35.5 \\
QuRating-Writing & 53.6 & 23.2 & \textbf{23.4} & 23.2 & 21.0 & 38.1 & 52.8 & 19.7 & 59.4 & 31.6 & 34.6 \\
QuRating-Edu &57.0 & 24.4 & 22.0 & 25.0 & 20.4 & \textbf{40.3} & 53.7 & 20.2 & 60.1 & 32.2 & 35.5 \\
FineWeb-Edu & 53.8 & 23.4 & 21.8 & 23.9 & 19.8 & 39.2 & 51.7 & 20.8 & 59.7 & 32.0 & 34.6 \\
DSIR-Book & 45.4 & 20.8 & 22.0 & 23.0 & 18.8 & 39.9 & \textbf{54.6} & 19.7 & 58.3 & 30.8 & 33.3 \\
DSIR-Wiki & 50.6 & 21.1 & 21.6 & 23.0 & 19.2 & 36.6 & 53.0 & 19.8 & 60.5 & 29.2 & 33.5 \\
\midrule
\textbf{Domain mixing methods} & & & & & & & & & & & \\
DOGE & 49.4 & 21.8 & 22.5 & 23.0 & 18.0 & 38.0 & 52.7 & 19.9 & 60.0 & 30.0 & 33.5 \\
DoReMi & 50.1 & 20.2 & 22.5 & 23.7 & 17.8 & 38.7 & 52.8 & 19.7 & 58.6 & 30.8 & 33.5 \\
DMLaw & 49.6 & 21.9 & 23.2 & 23.6 & 17.8 & 38.6 & 51.8 & 20.1 & 60.4 & 29.0 & 33.6 \\
RegMix &  50.0 &22.3& 22.1 	& 22.9 &	 18.8 &	 38.0 &	 52.8 &	 19.9 &	 58.9 &	 31.2 &33.7
 \\
\midrule
\textbf{Influence functions} & & & & & & & & & & & \\
MATES & 50.0 & 21.4 & 22.7 & 25.3 &	19.0 & 39.8 & 53.6 & \textbf{21.3} & 59.9 & 32.1 & 34.5 
 \\
\midrule
\textbf{Multi-actor Collaboration (Ours)} & \textbf{61.1} & \textbf{28.2} & 22.6 & \textbf{26.0} & \textbf{24.4} & 38.2 & 54.2 & 19.5 & 61.0 & 29.8 & \textbf{36.5} \\
\bottomrule
\end{tabular}
}
\end{table*}
\begin{table*}[!htb]
\centering
\caption{Table showing various selection methods and their three-shots performance. We report accuracy for all tasks, and \textbf{bold} the best result in each column. Abbreviations: O.B.QA = OpenbookQA W.G. = WinoGrande, C.S.QA = CommonSenseQA, Compreh. = Comprehensions
}\label{table:threeshot}
\vspace{1em}
\adjustbox{max width=0.9\textwidth}{
\begin{tabular}{l>{\centering\arraybackslash}m{0.085\textwidth}>{\centering\arraybackslash}m{0.085\textwidth}>{\centering\arraybackslash}m{0.085\textwidth}>{\centering\arraybackslash}m{0.085\textwidth}>{\centering\arraybackslash}m{0.085\textwidth}>{\centering\arraybackslash}m{0.085\textwidth}>{\centering\arraybackslash}m{0.085\textwidth}>{\centering\arraybackslash}m{0.085\textwidth}>{\centering\arraybackslash}m{0.085\textwidth}>{\centering\arraybackslash}m{0.085\textwidth}>{\centering\arraybackslash}m{0.085\textwidth}}
\toprule
 & \multicolumn{4}{c}{ Problem Solving }  & \multicolumn{4}{c}{ Commonsense Reasoning} & \multicolumn{2}{c}{ Reading Compreh.} & \\
\cmidrule(lr){2-5} \cmidrule(lr){6-9} \cmidrule(lr){10-11} 
 \multirow{1}{*}{\textbf{Selection Method}}& \textbf{ARC-E} & \textbf{ARC-C} & \textbf{MathQA} & \textbf{MMLU} & \textbf{O.B.QA} & \textbf{SIQA} & \textbf{W.G.} & \textbf{C.S.QA} & \textbf{BoolQ} & \textbf{RACE} & \textbf{Average} \\
\midrule
\textbf{Random sample} & & & & & & & & & & & \\
Uniform-30B & 54.6 & 23.0 & 22.1 & 24.9 & 18.8 & 40.3 & 52.9 & \textbf{21.5} & 53.0 & 29.8 & 34.1 \\
Uniform-60B & 58.8 & 25.5 & 23.0 & \textbf{27.2} & 20.0 & \textbf{41.8} & 53.6 & 19.6 & 56.9 & \textbf{32.7} & 35.9 \\
\midrule
\textbf{Perplexity-based data selection} & & & & & & & & & & & \\
PPL & 50.6 & 21.3 & 22.7 & 25.2 & 15.6 & 37.7 & 48.9 & 20.1 & 61.5 & 22.3 & 32.6 \\
\midrule
\textbf{Classifier-based data selection} & & & & & & & & & & & \\
QuRating-Facts & 59.5 & 25.7 & 22.6 & 25.9 & 19.8 & 40.2 & \textbf{54.6} & 19.2 & 60.8 & 24.8 & 35.3 \\
QuRating-Req & 59.3 & 25.9 & 22.7 & 26.1 & 19.6 & 39.7 & 53.7 & 20.5 & 58.5 & 22.7 & 34.9 \\
QuRating-Writing & 56.9 & 25.7 & \textbf{23.1} & 26.0 & 20.4 & 41.1 & 53.6 & 20.2 & 51.4 & 22.6 & 34.1 \\
QuRating-Edu & 60.8 & 26.5 & 22.5 & 26.7 & 20.2 & 41.4 & 54.6 & 20.6 & 55.5 & 22.7 & 35.1 \\
FineWeb-Edu & 56.2 & 25.7 & 22.3 & 26.2 & 20.6 & 40.1 & 50.5 & 19.7 & 56.6 & 31.4 & 34.9 \\
DSIR-Book & 48.7 & 21.0 & 22.6 & 25.6 & 18.6 & 42.5 & 53.7 & 19.5 & 57.9 & 22.9 & 33.3 \\
DSIR-Wiki & 53.2 & 22.4 & 22.6 & 25.3 & 17.6 & 37.1 & 52.7 & 21.4 & \textbf{61.6} & 24.2 & 33.8 \\
\midrule
\textbf{Domain mixing methods} & & & & & & & & & & & \\
DOGE & 52.4 & 21.9 & 22.4 & 27.0 & 17.4 & 39.9 & 52.0 & 18.2 & 57.8 & 29.8 & 33.9 \\
DoReMi & 53.2 & 21.4 & 22.2 & 24.7 & 18.2 & 38.4 & 50.9 & 20.6 & 59.7 & 31.1 & 34.0 \\
DMLaw & 51.5 & 21.4 & 22.4 & 25.2 & 18.2 & 39.0 & 50.7 & 19.4 & 52.6 & 29.8 & 33.0 \\
RegMix & 53.1 &	 22.1 	& 22.2 &	 25.4& 	 19.0 &	 39.1 &	 53.5 &	 18.4 &	 60.7 & 	 30.0 &	 34.4 
 \\
\midrule
\textbf{Influence functions} & & & & & & & & & & & \\
MATES & 52.6 & 21.8	& 22.6 & 26.7 &	20.4 & 40.9	& 53.7 & 19.7 &	57.6 & 31.8 & 34.8 
 \\
\midrule
\textbf{Multi-actor Collaboration (Ours)} & \textbf{65.8} & \textbf{31.5} & 23.0 & 26.6 & \textbf{24.6} & 39.9 & 54.1 & 20.1 & 60.4 & 30.5 & \textbf{37.7} \\
\bottomrule
\end{tabular}
}
\end{table*}

\begin{table*}[!htb]
\centering
\caption{Table showing various selection methods and their five-shots performance. We report accuracy for all tasks, and \textbf{bold} the best result in each column. Abbreviations: O.B.QA = OpenbookQA W.G. = WinoGrande, C.S.QA = CommonSenseQA, Compreh. = Comprehensions
}\label{table:fiveshot}
\vspace{1em}
\adjustbox{max width=0.9\textwidth}{
\begin{tabular}{l>{\centering\arraybackslash}m{0.085\textwidth}>{\centering\arraybackslash}m{0.085\textwidth}>{\centering\arraybackslash}m{0.085\textwidth}>{\centering\arraybackslash}m{0.085\textwidth}>{\centering\arraybackslash}m{0.085\textwidth}>{\centering\arraybackslash}m{0.085\textwidth}>{\centering\arraybackslash}m{0.085\textwidth}>{\centering\arraybackslash}m{0.085\textwidth}>{\centering\arraybackslash}m{0.085\textwidth}>{\centering\arraybackslash}m{0.085\textwidth}>{\centering\arraybackslash}m{0.085\textwidth}}
\toprule
 & \multicolumn{4}{c}{ Problem Solving }  & \multicolumn{4}{c}{ Commonsense Reasoning} & \multicolumn{2}{c}{ Reading Compreh.} & \\
\cmidrule(lr){2-5} \cmidrule(lr){6-9} \cmidrule(lr){10-11} 
 \multirow{1}{*}{\textbf{Selection Method}}& \textbf{ARC-E} & \textbf{ARC-C} & \textbf{MathQA} & \textbf{MMLU} & \textbf{O.B.QA} & \textbf{SIQA} & \textbf{W.G.} & \textbf{C.S.QA} & \textbf{BoolQ} & \textbf{RACE} & \textbf{Average} \\
\midrule
\textbf{Random sample} & & & & & & & & & & & \\
Uniform-30B & 54.5 & 21.9 & 22.4 & 25.6 & 19.2 & 39.7 & 54.2 & 19.7 & 53.2 & 30.8 & 34.1 \\
Uniform-60B & 59.1 & 26.0 & 22.4 & \textbf{26.9} & 21.6 & \textbf{42.1} & 54.3 & 21.0 & 55.7 & \textbf{32.4} & 36.2 \\
\midrule
\textbf{Perplexity-based data selection} & & & & & & & & & & & \\
PPL & 49.2 & 21.2 & 22.5 & 24.9 & 14.6 & 36.7 & 49.8 & 20.6 & 60.6 & 23.3 & 32.4 \\
\midrule
\textbf{Classifier-based data selection} & & & & & & & & & & & \\
QuRating-Facts & 60.5 & 25.4 & \textbf{23.4} & 26.4 & 20.2 & 40.3 & 51.9 & 19.0 & 58.0 & 23.3 & 34.8 \\
QuRating-Req & 59.9 & 26.4 & 22.8 & 25.6 & 21.8 & 40.1 & 53.7 & 19.6 & 56.9 & 22.3 & 34.9 \\
QuRating-Writing & 57.3 & 25.3 & 22.6 & 25.0 & 21.4 & 41.6 & 53.5 & 19.6 & 49.5 & 22.1 & 33.8 \\
QuRating-Edu & 60.8 & 26.5 & 22.5 & 26.5 & 20.2 & 41.4 & \textbf{54.6} & \textbf{21.1} & 55.5 & 22.7 & 35.2 \\
FineWeb-Edu & 56.6 & 24.9 & 22.6 & 25.8 & 19.8 & 39.4 & 51.2 & 19.7 & 55.9 & 30.9 & 34.7 \\
DSIR-Book & 49.7 & 21.1 & 22.1 & 25.6 & 19.8 & 41.7 & 54.1 & 18.3 & 55.6 & 22.9 & 33.1 \\
DSIR-Wiki & 53.6 & 22.3 & 23.0 & 25.3 & 17.6 & 36.7 & 52.2 & 20.4 & 60.2 & 22.6 & 33.4 \\
\midrule
\textbf{Domain mixing methods} & & & & & & & & & & & \\
DOGE & 53.0 & 21.8 & 22.0 & 26.3 & 17.2 & 40.1 & 51.7 & 18.8 & 58.5 & 30.1 & 33.9 \\
DoReMi & 52.7 & 22.2 & 22.4 & 25.5 & 16.2 & 39.3 & 51.9 & 20.9 & 60.0 & 31.0 & 34.2 \\
DMLaw & 52.4 & 21.4 & 23.0 & 25.7 & 17.2 & 39.2 & 50.6 & 19.2 & 51.4 & 29.9 & 33.0 \\
RegMix & 53.5 	& 24.0 & 21.2 & 25.0 & 19.6 & 41.0 & 53.2 & 19.0 & \textbf{61.3} & 30.2 & 34.8  \\
\midrule
\textbf{Influence functions} & & & & & & & & & & & \\
MATES & 53.6 &21.6 & 22.6 &	26.1 & 20.4 & 41.7 & 53.1 &	20.4 & 60.1 & 32.0 & 35.2 
 \\
\midrule
\textbf{Multi-actor Collaboration (Ours)} & \textbf{64.9} & \textbf{31.1} & 22.4 & 26.3 & \textbf{23.6} & 39.0 & 53.1 & 20.4 & 60.4 & 30.7 & \textbf{37.2} \\
\bottomrule
\end{tabular}
}
\end{table*}

We show the full results of all tasks in Table \ref{table:zershot}, Table \ref{table:threeshot} and Table \ref{table:fiveshot}. In analyzing the full experiment results, it is evident that our model consistently outperforms other methods across various tasks. Overall, for the zero-shot scenario, the classifier method outperforms the influence function in terms of average performance, while domain mixing yields the poorest results. Our method achieves an impressive average accuracy of 36.5, significantly surpassing the next best classifier, QuRating's series, which scores 35.5. This underscores the robustness of our approach, particularly in challenging problem-solving domains such as ARC-C, ARC-E, and MMLU, where we exceed competing models by considerable margins.

Our model demonstrates superior performance in the three-shot scenario, achieving an impressive average accuracy of 37.7, thereby maintaining its lead. Notably, we excel in the ARC-E and ARC-C benchmarks, attaining scores of 65.8 and 31.5, respectively, which highlights our model's effective utilization of few-shot learning. In contrast, the leading alternative methods underperform, particularly in more complex tasks such as MMLU and BoolQ.

In the five-shots evaluation, our model continues to demonstrate competitive performance, with scores reflecting a consistent trend of superiority across various domains, while other non-leading methods also maintain high levels. These results underscore our model's robust capacity to generalize across diverse question-answering tasks, affirming its advantages over conventional classifiers and highlighting its potential for practical applications in real-world scenarios.

\subsection{Implementation Details of our Methods}
\label{sec:app:theo}
To further refine the model's performance, we calculate rewards for each sampled data point by approximating the rewards using influence functions, as shown in Equation \ref{eq:influence}. Following \cite{dsdm}, we choose LAMBADA~\cite{paperno2016lambada}, SQuAD~\cite{rajpurkar2016squad} and Jeopardy~\cite{tunguz2019jeopardy} as reference tasks. We followed methods provided in \cite{dsdm}, \cite{less} and \cite{trak} to calculate the Hessian and the gradients in the influence functions. In our implementation, we project gradients into an 8,192-dimensional space for both the validation and training datasets. To optimize the gradient computation process, we divide each data category into eight slices, thereby enabling parallel computation across eight GPUs. Each slice contains 1,250 data points. After calculating gradients for each slice, the results are concatenated in their original sequence to ensure data integrity. This slicing strategy not only accelerates the processing by utilizing GPU parallelism but also maintains consistency in gradient calculation. Additionally, for the validation datasets, we uniformly sample 500 data points to ensure a balanced evaluation procedure. All prompts across the datasets are carefully aligned to maintain task coherence, a crucial factor in multi-task learning scenarios. Furthermore, we implement a sliding window of 1,024 tokens with a 256-token overlap to ensure consistent tokenization across the entire dataset. This sliding window technique efficiently extracts a maximum of 1,024 tokens from each data point, ensuring uniform encoding across different datasets and tasks, thus improving the overall consistency and reliability of the data processing pipeline.
\subsection{Details of Influence Changes During Different Pretraining Stages}
\label{sec:app:influence}
\begin{figure*}[!htb]
    \vspace{-0.5em}
    \centering
    \includegraphics[width=0.9\textwidth]{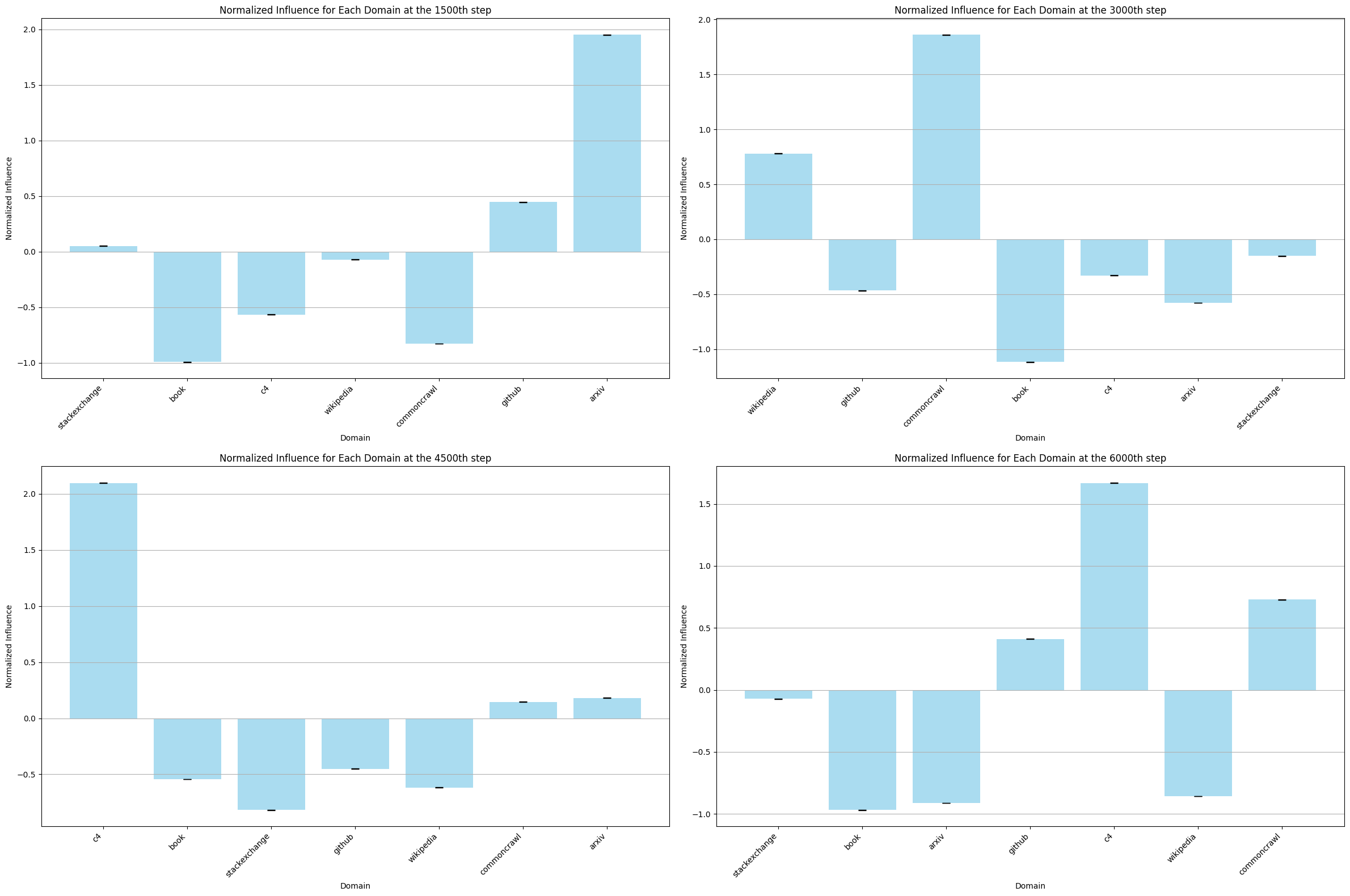}
    \caption{We present the normalized influence for each domain across various training steps.}
    \label{fig:topic_distribution_domain}
    \vspace{-1em}
\end{figure*}

\begin{figure*}[!htb]
    \centering
    \includegraphics[width=0.9\textwidth]{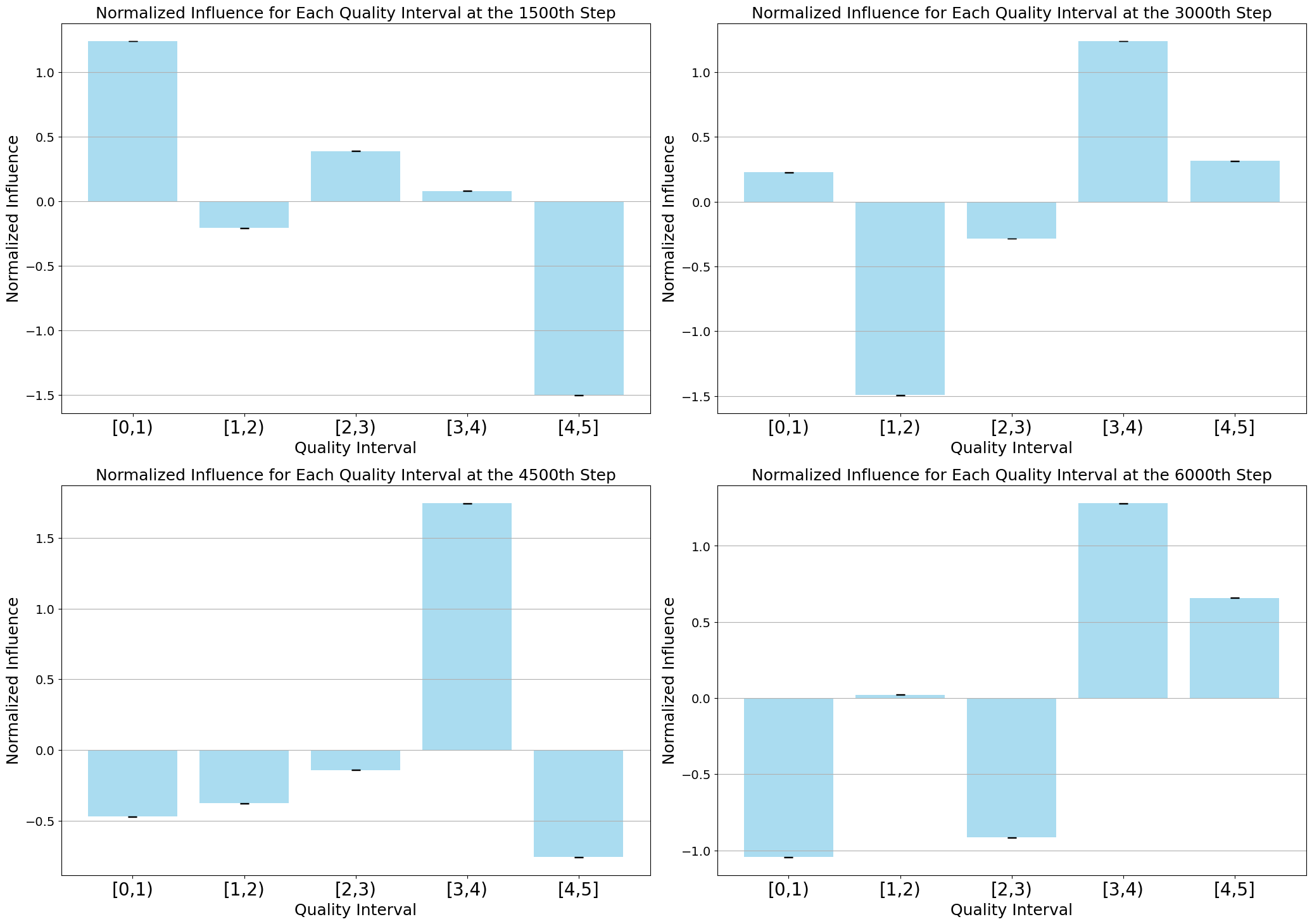}
    \caption{We present the normalized influence for each quality interval across various training steps.}
    \label{fig:topic_distribution_quality}
    \vspace{-0.5em}
\end{figure*}

\begin{figure*}[!htb]
    \centering
    \includegraphics[width=0.9\textwidth]{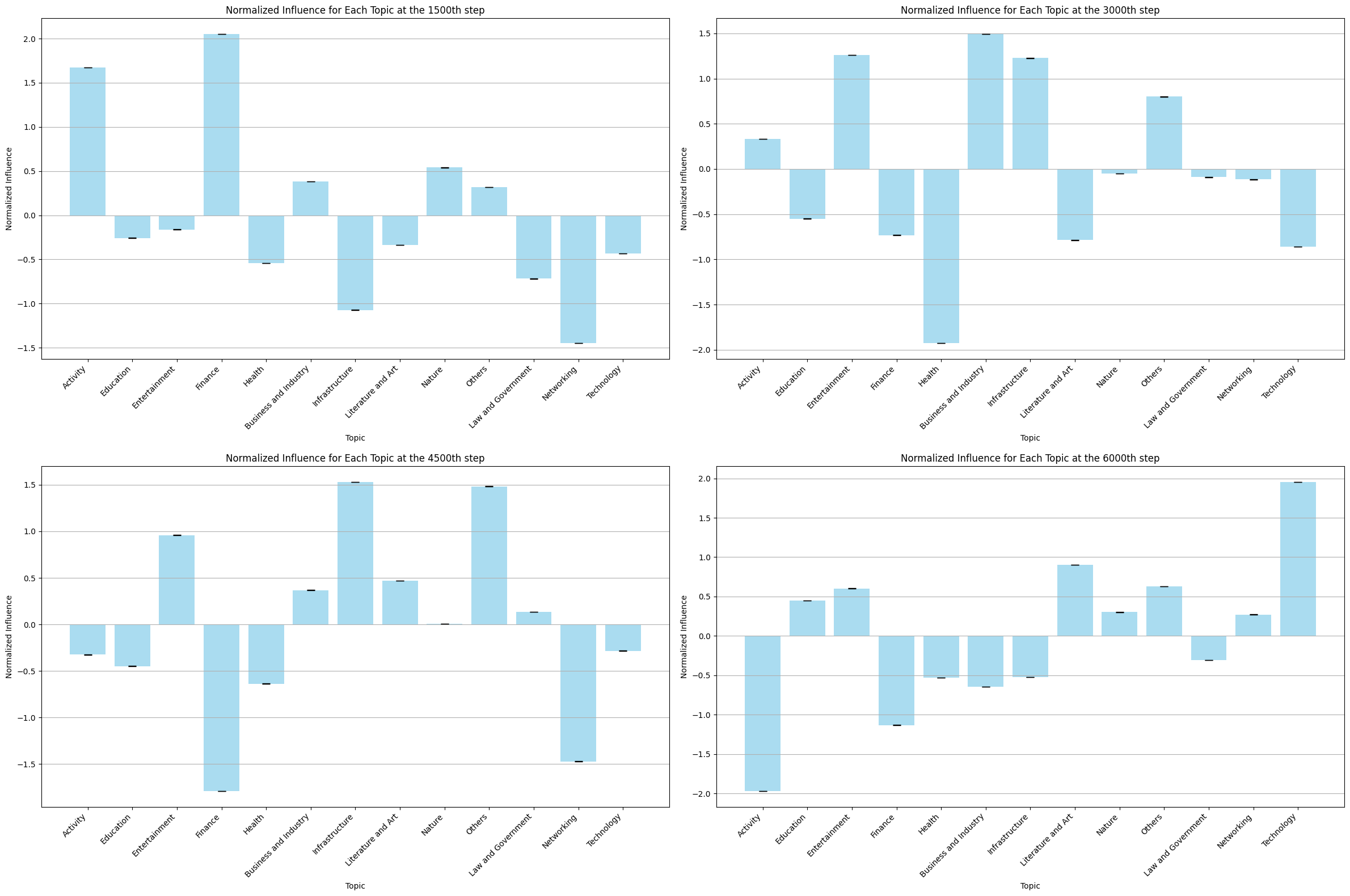}
    \caption{We present the normalized influence for each topic across various training steps.}
    \label{fig:topic_distribution_topic}
    \vspace{-0.5em}
\end{figure*}

We present the details of influence change during the pretraining process for domain (Figure \ref{fig:topic_distribution_domain}), quality intervals (Figure \ref{fig:topic_distribution_quality}) and topic (Figure \ref{fig:topic_distribution_topic}).

\subsection{Data Distribution Analysis of the SlimPajama Dataset}
\begin{figure*}[!htb]
\centering
\includegraphics[width=0.8\textwidth]{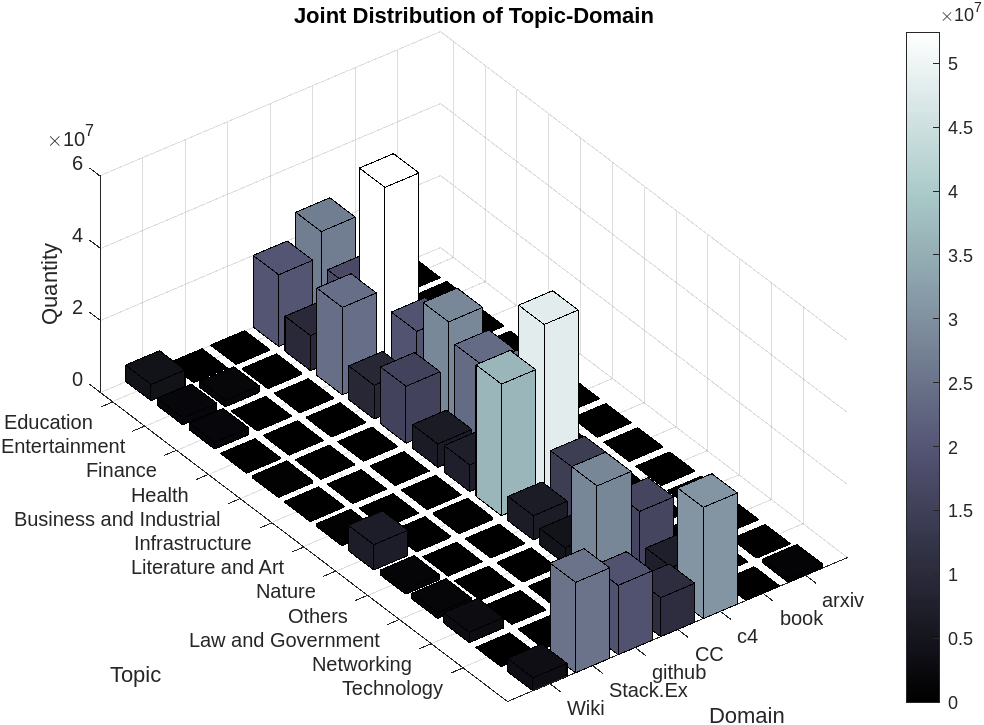}
    \caption{The illustration of the joint distribution of topics and domains.}
    \label{fig:joint_distribution_1}
\end{figure*}

\begin{figure*}[!htb]
    \centering
    
    \includegraphics[width=0.8\textwidth]{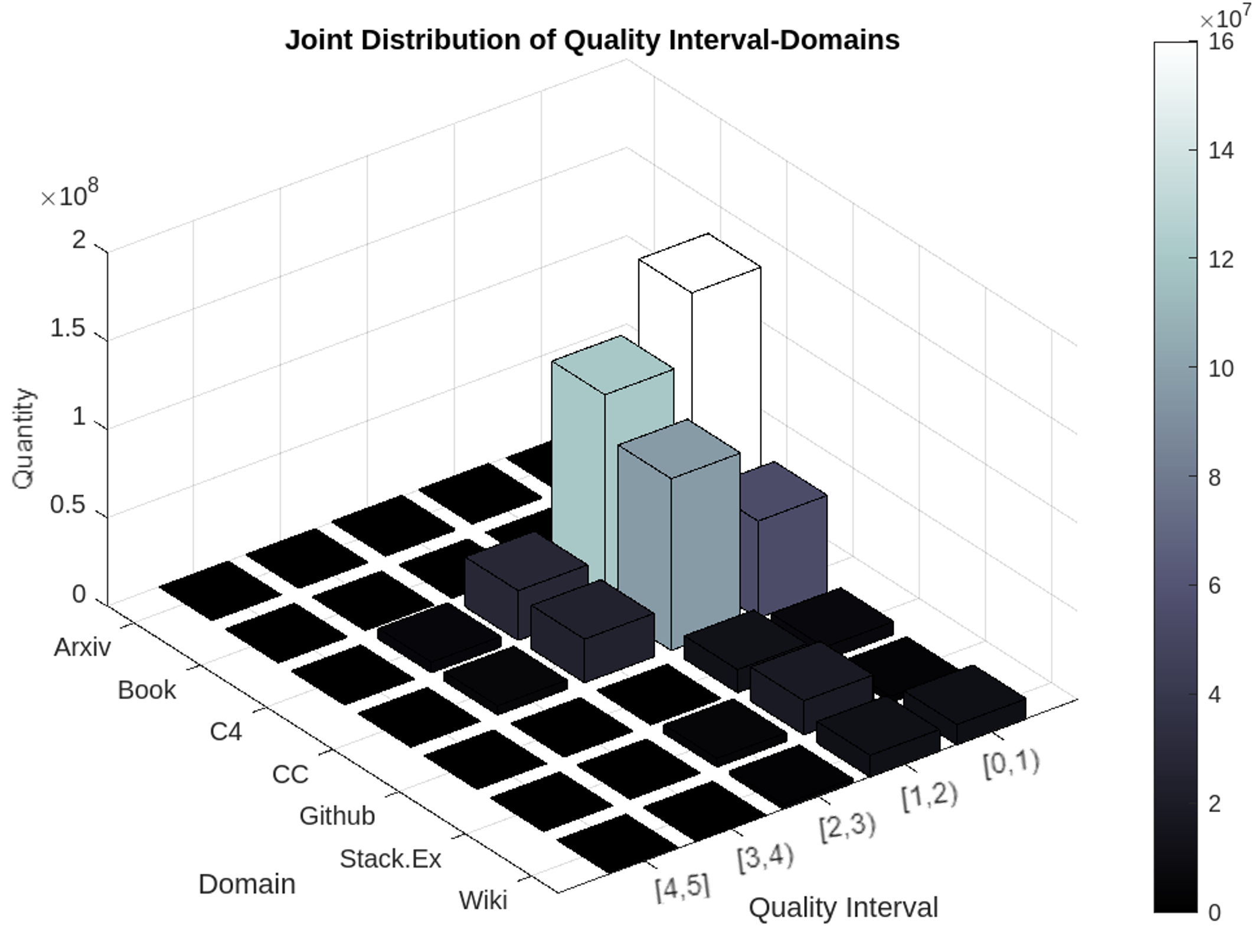}
    \caption{The illustration of the joint distribution of quality intervals and domains.}
    \label{fig:joint_distribution_2}
    
    \includegraphics[width=0.8\textwidth]{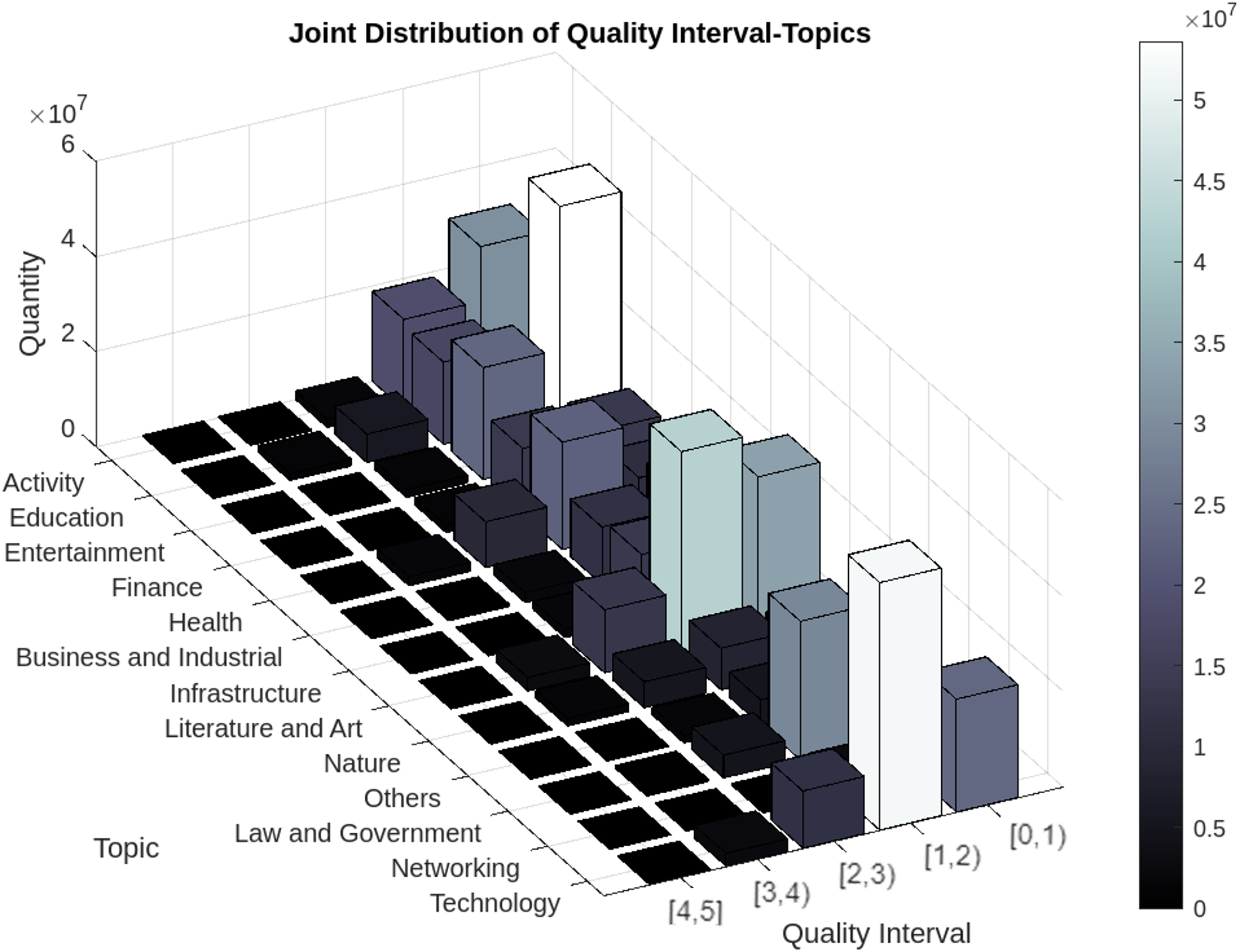}
    \caption{The illustration of the joint distribution of quality intervals and topics.}
    \label{fig:joint_distribution_3}
\end{figure*}

We finally present the data distribution analysis of the SlimPajama dataset from three dimensions: topic, domain and quality intervals, as Figure \ref{fig:joint_distribution_1} to Figure \ref{fig:joint_distribution_3} shows.

\end{document}